\documentclass[letterpaper]{article}
\usepackage[preprint]{aaai2027}
\usepackage[hyphens]{url}
\usepackage{graphicx}
\usepackage{natbib}
\usepackage{caption}
\usepackage{amsmath}
\usepackage{amssymb}
\usepackage{amsthm}
\usepackage{booktabs}

\theoremstyle{plain}
\newtheorem{theorem}{Theorem}
\newtheorem{proposition}{Proposition}
\newtheorem{corollary}{Corollary}

\newcommand{\externalhref}[2]{\leavevmode\pdfstartlink attr{/Border [0 0 0]} user{/Subtype /Link /A << /S /URI /URI (\pdfescapestring{#1}) >>}#2\pdfendlink}

\title{Beyond Prediction Residuals: Conformal Regret Calibration for Robust Contextual Linear Optimization}

\author{%
  Lingjie Zhao\textsuperscript{\rm 1},
  Hansheng Jiang\textsuperscript{\rm 2},
  Wei Qi\textsuperscript{\rm 1}
}

\affiliations{%
  \textsuperscript{\rm 1}Department of Industrial Engineering,
  Tsinghua University\\
  \textsuperscript{\rm 2}Rotman School of Management,
  University of Toronto
}

\begin{document}
\maketitle

\begin{abstract}
Predict-then-optimize systems convert contextual forecasts into decisions, yet conformal robust optimization typically calibrates prediction residuals rather than downstream loss. For linear optimization, this parameter-space geometry can be badly misaligned with decision risk: large objective perturbations may preserve the optimizer, while small boundary-crossing perturbations may cause substantial regret. We introduce \emph{Conformal Regret Sets}, a post-hoc, model-agnostic framework that conformalizes an optimization-derived regret surrogate and uses its calibrated level set for minimax-regret refinement. We first show that directly conformalizing exact regret, although sufficient to certify the nominal predict-then-optimize decision, yields a degenerate robust problem: absent an exogenous bounded support, every alternative decision has unbounded worst-case regret. We therefore use SPO+, a pointwise upper bound on exact regret, as the conformal score, obtaining distribution-free finite-sample marginal regret guarantees under exchangeability. For contextual linear programs, the resulting set decomposes into polyhedral pieces aligned with the normal fan of the feasible polytope. A vertex is active if and only if its predicted suboptimality gap is at most half the calibrated threshold $\eta$, enabling local, output-sensitive generation of all relevant pieces. Moreover, if $\Gamma$ is the predicted gap to the closest competing vertex, the nominal and refined decisions admit the sharper worst-case-regret certificate $(\eta-2\Gamma)_+$, with an exact early exit when $\eta\leq 2\Gamma$. Experiments on contextual fractional knapsack validate the geometry and certificates: continuous refinement reduces the $0.95$-quantile and $\operatorname{CVaR}_{0.95}$ of regret by $9.1\%$ and $13.1\%$ relative to nominal PTO, while vertex-restricted refinement changes only $10.0\%$ of nominal decisions.
\end{abstract}

\section{Introduction}
\label{sec:intro}

Predictive models are increasingly embedded in downstream optimization systems. In the standard predict-then-optimize (PTO) pipeline, observable features are mapped to uncertain problem parameters, and the resulting point prediction is inserted into a deterministic optimization problem. This modular paradigm is natural for contextual linear programs (LPs) arising in routing, allocation, and portfolio selection \citep{bertsimas1998introduction,bertsimas2020predictive}. Yet a point forecast alone offers no protection against prediction error, and optimizing an estimated objective can amplify modest errors into substantial post-decision loss \citep{smith2006optimizer}.

A common remedy is to replace the point prediction with a data-driven uncertainty set and solve a robust optimization problem. Conformal prediction (CP) is attractive for this purpose because it is model-agnostic and provides distribution-free, finite-sample marginal coverage under exchangeability \citep{angelopoulos2023conformal}. Existing conformal approaches to contextual optimization typically calibrate residual scores, such as coordinatewise errors or norm distances between predicted and realized objective vectors \citep{sun2023predict,patel2024conformal}. The resulting boxes, ellipsoids, or norm balls are statistically valid in parameter space, but need not be aligned with downstream decision risk.

For an LP, residual coverage is not decision calibration. The objective space is partitioned by the normal fan of the feasible polytope: all costs in one cone select the same optimal vertex. A large perturbation can remain within a cone and incur no decision loss, whereas a smaller perturbation can cross a nearby boundary, change the optimizer, and cause substantial regret. Some directions also shift every feasible objective value equally and are therefore irrelevant to both the optimizer and pairwise regret. Residual-based scores do not distinguish these decision-invariant directions from decision-sensitive ones. This leads to the following question: \emph{Can conformal uncertainty instead be calibrated directly in units of downstream regret while still supporting nontrivial robust re-optimization?}

Conformal validity alone is not the obstacle: the score sublevel set must also support meaningful comparison of alternative actions. The most direct score is exact decision regret, or the Smart Predict-then-Optimize (SPO) loss \citep{elmachtoub2022smart}. Although its conformal quantile immediately certifies the nominal PTO decision, we prove that, without an independently imposed bounded objective support, its sublevel set is degenerate for re-optimization: every alternative decision has infinite worst-case regret. Thus, a score that is ideal for certifying one fixed action can be unusable for comparing actions.

We therefore introduce \emph{Conformal Regret Sets} based on SPO+, a pointwise upper bound on exact regret. Split-conformal calibration produces a threshold $\eta$ in objective-value units, yielding a finite-sample marginal regret guarantee and an objective-space set for minimax-regret refinement. The construction is post hoc and compatible with any black-box predictor. For contextual LPs, the SPO+ level set has an exact normal-fan structure that makes this otherwise nonconvex robust problem analyzable and amenable to output-sensitive solution methods.

Our contributions are threefold.
\begin{itemize}
    \item \textbf{Regret calibration and the need for a robustifiable score.}
    We formalize conformal calibration with optimization regret, establish the degeneracy of exact-SPO sublevel sets for robust refinement, and show that SPO+ calibration provides a distribution-free finite-sample guarantee for the regret of the nominal PTO decision.

    \item \textbf{Exact normal-fan geometry and local active-set generation.}
    We derive a closed polyhedral representation of the SPO+ conformal regret set, exact whenever the hindsight optimizer is unique, and show that its unbounded directions are decision-invariant. A vertex generates a nonempty piece if and only if its predicted suboptimality gap is at most $\eta/2$. These active vertices form a connected subgraph of the polytope's one-skeleton, enabling output-sensitive generation from the nominal solution rather than enumeration of all vertices.

    \item \textbf{Margin-certified minimax-regret refinement.}
    Let $\Gamma$ be the predicted gap between the nominal optimizer and its closest competing vertex. We prove that the certified worst-case regret of the nominal, vertex-restricted, and continuous refinements is at most $(\eta-2\Gamma)_+$. With $\eta=\eta_\alpha$, the same bound controls realized regret with probability at least $1-\alpha$. Hence $\eta\leq2\Gamma$ gives a zero worst-case-regret certificate over the calibrated set and an exact early-exit test. Otherwise, adjacency-based generation and a fixed-piece LP oracle evaluate the regret adversary without enumerating all comparator vertices; once the active pieces are generated, the continuous refinement admits a finite LP formulation.
\end{itemize}

\section{Related Work}

\textbf{Conformal prediction for optimization.}
CP provides distribution-free prediction regions under exchangeability \citep{angelopoulos2023conformal}, and a growing literature uses these regions as uncertainty sets for contextual robust optimization \citep{zhao2026conformal}. Early methods primarily calibrate parameter-space errors, including ellipsoidal sets \citep{pmlr-v152-johnstone21a}, residual-based sets for contextual LPs \citep{sun2023predict}, and generative, potentially nonconvex regions \citep{patel2024conformal}. Conformal ideas have also been combined with inverse optimization and adherence-aware prescription \citep{chan2024conformal}. More broadly, conformal risk control extends conformal guarantees beyond miscoverage to generic monotone losses \citep{angelopoulos2024risk}.

\textbf{Decision-aware conformal score design.} Beyond residual calibration, recent work makes uncertainty quantification for optimization more decision-aware. End-to-end conditional robust optimization trains predictive and uncertainty models using downstream decision loss \citep{chenreddy2024end}, while conformal methods learn calibrated convex or polyhedral set geometry from the induced robust loss \citep{yeh2025end,chen2026polyhedral}. In particular, \citet{chen2026polyhedral} learn geometry within a parameterized polyhedral family and then recalibrate on independent data. Utility-directed CP incorporates a user-specified downstream utility into conformal score design \citep{cortesgomez2025utility}; \citet{hu2026conformal} design sets under explicit robustness constraints, and \citet{zhou2026inverse} calibrate downstream robustness levels. CREDO combines conformal and inverse-optimization geometry to audit the probability that a fixed candidate decision is suboptimal \citep{zhou2026credo}. Our construction derives set geometry directly from the optimization-based SPO+ score. The resulting nonconvex, normal-fan-aligned sets admit exact active-piece screening, margin-adaptive certificates, and certified minimax-regret refinement for contextual LPs.

\textbf{Smart predict-then-optimize and decision-focused learning.}
Decision-focused learning (DFL) addresses the mismatch between prediction accuracy and downstream performance by differentiating through, or otherwise exploiting, the optimization problem during training \citep{donti2017task,wilder2019melding,mandi2022decision,gupta2024decision}. The Smart PTO framework uses decision regret as the training objective and introduces SPO+ as a tractable convex surrogate for linear optimization \citep{elmachtoub2022smart}. \citet{liu2021risk} study surrogate-risk calibration, relating excess SPO+ risk to excess expected SPO risk to support predictor learning. Subsequent work develops optimization-aware losses and training procedures across linear, combinatorial, probabilistic, and causal settings \citep{qi2025integrated,berden2025solverfree,zhang2025bilevel}; robust-constraint-aware training is considered through the SPO-RC loss \citep{im2025smart}. Our calibration is post hoc: we apply split-conformal calibration to SPO+ scores from a fixed predictor, obtaining finite-sample marginal coverage and regret guarantees for minimax-regret refinement.

\section{Problem Setup and Regret Calibration}
\label{sec:problem-setup}
\subsection{Contextual Linear Optimization and Decision Regret}
\label{sec:contextual_lp}
Consider a linear program (LP) that takes the following standard form:
\begin{align}
    \min &\ c^\top x,\\
    \text{s.t.}&\ Ax=b, x\ge0,
\end{align}
where $x \in \mathbb{R}^n$ is the decision variable, $c \in \mathbb{R}^n$ is the uncertain parameter, and $A \in \mathbb{R}^{m \times n}$ and $b \in \mathbb{R}^m$ ($m \le n$) define the deterministic system constraints. We assume throughout that $\mathcal{X}:=\{x\in\mathbb{R}^n:Ax=b,\ x\geq0\}$ is a nonempty compact polytope. Let $\mathcal{V}:=\operatorname{ext}(\mathcal X)=\{v_1,\ldots,v_K\}$ be its complete set of distinct geometric vertices, and write $[K]:=\{1,\ldots,K\}$. For any realized cost vector $c$, let $x^*(c)\in\arg\min_{x \in \mathcal{X}} c^\top x$ be a fixed deterministic measurable vertex-valued optimal selector.
Throughout the main text, we assume that every cost vector under consideration induces a unique optimizer, so the selector is immaterial there; tie-boundary extensions are treated in the appendix.

For a feasible decision $x\in \mathcal{X}$ and a realized objective vector $c$, define its decision regret as
\begin{equation}
    \operatorname{Reg}(x,c):=c^\top x-c^\top x^*(c).
    \label{eq:decision-regret}
\end{equation}
It measures the excess realized cost of $x$ relative to the hindsight optimum and is therefore nonnegative.

In the Predict-Then-Optimize paradigm, a predictor $\hat{f}: \mathbb{R}^d\rightarrow\mathbb{R}^n$ maps observable features $z$ to a point estimate $\hat{c}$ of the uncertain parameter $c$. The PTO simply prescribes $x^*(\hat{c})$, but the fragility of decisions based on point forecasts motivates robust methods, i.e., replacing $\hat{c}$ with an uncertainty set and optimizing against its worst-case realization.

\subsection{Split Conformal Calibration}
\label{sec:split_cp}
To quantify predictive uncertainty, we leverage split conformal prediction \citep{angelopoulos2023conformal}. Let $\mathcal{D}_{cal} = \{(z_i, c_i)\}_{i=1}^{N}$ be a held-out calibration sample set, and define a conformal score function $s(z, c)$ measuring the discrepancy between the prediction $\hat{f}(z)$ and the realization $c$. Evaluating this score on $\mathcal{D}_{cal}$ yields empirical scores $\mathcal{S} = \{s_i\}_{i=1}^N$ and the order statistics $s_{(1)}\leq\cdots\leq s_{(N)}$. For a target miscoverage level $\alpha \in (0,1)$, set 
\begin{equation}
    q_\alpha:=\left\lceil (N+1)(1-\alpha)\right\rceil,
    \qquad\eta_\alpha:=s_{(q_\alpha)},
    \label{eq:conformal-quantile}
\end{equation}
with $s_{(N+1)}=+\infty$. For any new test instance $z$, the conformal uncertainty set is constructed as the level set
\begin{equation}
    \mathcal{U}_\alpha(z) = \left\{ c \in \mathbb{R}^n : s(z, c) \le \eta_\alpha \right\}.
\end{equation}

Under the standard exchangeability assumption, the foundational result in conformal prediction yields a distribution-free finite-sample marginal coverage guarantee,
\[
\mathbb{P}(c_{\mathrm{test}}\in\mathcal{U}_\alpha(z_{\mathrm{test}}))\ge1-\alpha.
\]
The choice of score determines the geometry and downstream meaning of the resulting set. We next specialize the score to decision-aware optimization settings.

\subsection{SPO+ Conformal Regret Set and Optimization}
\label{sec:conformal_regret_set}

Existing conformal robust optimization methods typically calibrate generic parameter-space sets, such as norm balls, boxes, ellipsoids, and budgeted sets. Although these sets admit well-established deterministic reformulations, they often ignore downstream decision geometry. 
This mismatch motivates constructing uncertainty sets by directly calibrating decision regret. For the nominal PTO decision, the natural score is its realized regret, namely the exact SPO loss \citep{elmachtoub2022smart}:
\begin{equation}
    s_{\text{SPO}}(c,\hat{c}) = c^\top x^*(\hat{c})-c^\top x^*(c).
\end{equation}

For any threshold $\eta\ge0$, define the SPO uncertainty set as
\begin{equation}
    \mathcal{U}_{\mathrm{SPO}}(\hat c;\eta):=
    \left\{c\in\mathbb{R}^{n}:s_{\mathrm{SPO}}(c,\hat c)\leq\eta\right\}.
    \label{eq:spo_uncertainty_set}
\end{equation}
By construction, every $c\in\mathcal{U}_{\mathrm{SPO}}(\hat c;\eta)$ certifies that the nominal PTO decision incurs regret at most $\eta$.
However, this fixed-decision certificate does not necessarily provide a useful uncertainty representation for selecting alternative prescriptions: as the following proposition shows, regret-robust refinement over the exact-SPO uncertainty sets is degenerate.

\begin{proposition}[Degeneracy of exact-SPO refinement]
\label{prop:spo_robustification_degeneracy}
Suppose that $\mathcal X$ contains at least two distinct points and that
$\hat x:=x^*(\hat c)$ is a vertex of $\mathcal X$. For every finite
$\eta\geq0$,
\begin{equation*}
    \sup_{c\in\mathcal U_{\mathrm{SPO}}(\hat c;\eta)}
    \operatorname{Reg}(\hat x,c)
    \leq \eta,
\end{equation*}
whereas, for every $x\in\mathcal X\setminus\{\hat x\}$,
\begin{equation*}
    \sup_{c\in\mathcal U_{\mathrm{SPO}}(\hat c;\eta)}
    \operatorname{Reg}(x,c)
    =+\infty.
\end{equation*}
Consequently, $\hat x$ is the unique minimizer of the corresponding
worst-case regret problem.
\end{proposition}

Proposition~\ref{prop:spo_robustification_degeneracy} shows that exact SPO is suitable for certifying the nominal PTO decision but cannot support nontrivial refinement: every alternative prescription has infinite worst-case regret over its sublevel set. An independently justified compact objective support can restore boundedness, but this boundedness is supplied externally rather than by the exact-SPO score.
Detailed proofs and a comprehensive treatment of the theoretical extensions are provided in the appendix.

This motivates using the SPO+ function for robustification. Unlike the action-specific exact-SPO certificate, SPO+ retains comparator-wise information across alternative decisions. For a realized parameter $c$ and prediction $\hat c$, define
\begin{equation}
    s_{\text{SPO+}}(c, \hat{c}) = \max_{x \in \mathcal{X}} \{ (c - 2\hat{c})^\top x \} + (2\hat{c} - c)^\top x^*(c),
\end{equation}
which pointwise bounds the exact regret (SPO function).
Applying the split-conformal procedure in Section~\ref{sec:split_cp} to the SPO+ scores yields the threshold $\eta_\alpha$ and the conformal regret set
\begin{equation}
    \mathcal{U}_{\text{SPO+}}(\hat{c};\eta_\alpha) = \{ c \in \mathbb{R}^n : s_{\text{SPO+}}(c, \hat{c}) \le \eta_\alpha \}.
\end{equation}
As formalized below, this construction translates conformal validity into a finite-sample guarantee on the regret of the nominal PTO decision. 

\begin{theorem}[Finite-sample regret validity]
\label{thm:spo-plus-validity}
Let $\hat c_{\mathrm{test}}:=\hat f(z_{\mathrm{test}})$. If the calibration and test observations are exchangeable and the predictor $\hat{f}$ is trained on data independent of the joint calibration–test sample, then
\begin{equation}
    \mathbb P\!\left\{
        c_{\mathrm{test}}\in \mathcal U_{SPO+}
        \bigl(\hat c_{\mathrm{test}};\eta_\alpha\bigr)
    \right\}
    \geq 1-\alpha.
    \label{eq:spo-plus-set-validity}
\end{equation}
Moreover, the nominal PTO decision satisfies
\begin{equation}
    \mathbb{P}\left( c_{\mathrm{test}}^\top x^*(\hat{c}_\mathrm{test}) - c_\mathrm{test}^\top x^*(c_\mathrm{test}) \le \eta_\alpha \right) \ge 1-\alpha.
\end{equation}
\end{theorem}
Theorem~\ref{thm:spo-plus-validity} provides an interpretable operational certificate by bounding the regret of a prescribed action before deployment. In conventional robust optimization, conservatism is controlled through geometric parameters, such as uncertainty-set radii or budgets, whose implications for decision quality may be difficult to interpret. In contrast, our framework links the coverage level $1-\alpha$ directly to a regret threshold $\eta_\alpha$, allowing practitioners to assess the trade-off between statistical reliability and tolerated suboptimality.

\paragraph{Downstream regret refinement.}
The conformal set can also be used to select a decision that is less exposed to adverse objective realizations than the nominal PTO prescription. For any $x\in \mathcal{X}$ and a generic threshold $\eta\geq 0$, define its worst-case regret over this set as
\begin{equation}
    \rho_\eta(x;\hat c):=\sup_{c\in\mathcal{U}_{SPO+}(\hat c;\eta)}\operatorname{Reg}(x,c).
    \label{eq:worst-case-regret-definition}
\end{equation}
Given a nonempty compact admissible decision domain $\mathcal D\subseteq \mathcal{X}$, the corresponding regret-robust refinement is
\begin{equation}
    x_{\mathcal D}^{\mathrm{RR}}(\hat c)\in
    \arg\min_{x\in\mathcal D} \rho_{\eta_\alpha}(x;\hat c).
    \label{eq:generic-regret-refinement}
\end{equation}
Later, we consider the two principal choices $\mathcal D=\mathcal{V}$, which restricts the refinement to feasible vertices, and $\mathcal D=\mathcal{X}$, which permits any feasible point.

\section{Geometry and Regret Refinement}
\label{sec:geometry_optimization}

Fix a prediction $\hat c$ and a threshold $\eta\geq0$, and write $\rho_\eta(x):=\rho_\eta(x;\hat c)$. Because the SPO+ conformal regret set is generally nonconvex, we first exploit its geometry rather than optimize over it monolithically. We first derive a vertex-indexed polyhedral decomposition and establish finite worst-case regret on each piece. We then characterize and locally generate the nonempty pieces, with certified truncation based on predicted gaps. Finally, we develop margin-based certificates and computational formulations for regret refinement without requiring a complete vertex description.

\subsection{Polyhedral Decomposition of Set Geometry}
\label{sec:decomposition}
Within a region where a fixed vertex $v_k$ is hindsight optimal, the SPO+ score is convex and piecewise linear in $c$, so its sublevel set is polyhedral. The union of these regional sets need not be convex. Under the main-text unique-optimizer assumption, this yields the following exact decomposition; the tie-safe extension is given in the appendix.

\begin{theorem}[Polyhedral Decomposition]
\label{thm:decomposition}
At any cost vector $c$ with unique optimizer $v_k$,
\[
c\in\mathcal U_{\mathrm{SPO+}}(\hat c;\eta)
\quad\Longleftrightarrow\quad c\in P_k,
\]
where
\begin{equation}\label{eq:poly_decompose}
\begin{aligned}
    P_k = \Bigl\{
    c \in \mathbb{R}^n :\;&
    c^\top(v_j-v_k) \le \eta+ 2\hat{c}^\top(v_j-v_k),\\
    &
    c^\top(v_k-v_j) \le 0,\ \forall j \in \{1,\dots,K\}
    \Bigr\}.
\end{aligned}
\end{equation}
\end{theorem}

Although $\bigcup_{k=1}^K P_k$ additionally includes cost vectors with tied optimizers, it is the closure of the unique-optimizer portion of $\mathcal U_{\mathrm{SPO+}}(\hat c;\eta)$ and, by continuity of regret, yields the same worst-case regret values. The second group of inequalities makes $v_k$ optimal under $c$, while the first imposes the SPO+ threshold conditional on $v_k$ being hindsight optimal. Hence, on $P_k$, $\operatorname{Reg}(x,c)=c^\top(x-v_k)$, and worst-case regret can be evaluated piecewise over the union.

An unbounded piece $P_k$ can still induce finite worst-case regret. The following result characterizes its recession cone $\operatorname{rec}(P_k)$ and the worst-case regret $\phi_k(x)$ of a feasible decision $x$ over that piece.

\begin{proposition}[Finite regret over each piece]
\label{prop:finite_piece_regret}
For every nonempty piece $P_k$,
\begin{equation}
    \operatorname{rec}(P_k)=\operatorname{span}(\mathcal X-\mathcal X)^\perp.
    \label{eq:piece_recession}
\end{equation}
Consequently, for every $x\in\mathcal X$,
\begin{equation}
    \phi_k(x):=\max_{c\in P_k}c^\top(x-v_k)
    \label{eq:piece_regret}
\end{equation}
is finite and attained.
\end{proposition}

Proposition~\ref{prop:finite_piece_regret} shows that the ambient unboundedness of $P_k$ is harmless for regret evaluation: its recession directions are decision invariant, so every active piece induces a finite and attained adversarial problem. Consequently, worst-case regret over the nonconvex set can be evaluated piecewise over the active pieces.

Let
\begin{equation}
    \mathcal{A}:=\{k\in[K]:P_k\neq\varnothing\}
    \label{eq:active_set}
\end{equation}
denote the active-piece index set. It follows that, for every
$x\in\mathcal{X}$,
\begin{equation}
    \rho_\eta(x)=
    \max_{k\in\mathcal{A}}\phi_k(x).
    \label{eq:piecewise_regret_representation}
\end{equation}
The decomposition is exact but may still be computationally large: $K$ can be exponential, and each $P_k$ in \eqref{eq:poly_decompose} contains constraints indexed by all comparator vertices. The next subsection addresses the first issue through active-piece screening and local generation, while Section~\ref{sec:certified_refinement} addresses the second through a constraint-based fixed-piece problem reformulation (oracle).

\subsection{Active-Piece Screening and Generation}
\label{sec:active_piece_generation}

We first address the number of potential pieces. For each $v_k\in\mathcal V$, define its predicted gap relative to $\hat v=x^*(\hat c)$ as
\begin{equation}
    \Delta_k:=\hat c^\top(v_k-\hat v)\geq 0.
    \label{eq:predicted_vertex_gap}
\end{equation}
The following result gives an exact characterization of the nonempty polyhedral pieces.

\begin{theorem}[Exact active-piece screening]
\label{thm:screening}
For every vertex $v_k\in\mathcal{V}$ and every finite threshold $\eta\geq0$,
\begin{equation}
    P_k\neq\emptyset
    \quad\Longleftrightarrow\quad
    \Delta_k\leq\frac{\eta}{2}.
    \label{eq:exact_active_screen}
\end{equation}
\end{theorem}

Therefore,
\begin{equation}
    \mathcal A= \left\{k\in[K]:\Delta_k\leq\frac{\eta}{2}\right\}.
    \label{eq:active_piece_set}
\end{equation}
The relevant complexity is determined by the number of $\eta/2$-near-optimal vertices rather than by $|\mathcal V|$. It may be much smaller when the nominal optimizer is well separated, although it can remain exponential in the worst case.

Screening alone would still require evaluating $\Delta_k$ for every vertex. We next show that the active vertices can be generated locally. Let $\mathcal G_{\mathcal X}$ denote the one-skeleton of $\mathcal X$, and for any $\tau\geq0$ define
\begin{equation}
    \mathcal{V}_{\tau}:=
    \left\{v\in\mathcal{V}:\hat c^\top(v-\hat v)\leq\tau\right\}.
    \label{eq:vertex_sublevel_set}
\end{equation}

\begin{proposition}[Connectivity of predicted near-optimal vertices]
\label{prop:active_connectivity}
For every $\tau\geq0$, the subgraph of $\mathcal{G}_{\mathcal{X}}$ induced by $\mathcal{V}_{\tau}$ is connected. More precisely, for every $v\in\mathcal{V}_{\tau}$, there exists an edge path $\hat v=v^{0},v^{1},\ldots,v^{L}=v$ such that
\begin{equation}
    \hat c^\top v^{0} \leq \hat c^\top v^{1} \leq\cdots\leq \hat c^\top v^{L},
    \label{eq:monotone_vertex_path}
\end{equation}
and every vertex on the path belongs to $\mathcal{V}_{\tau}$.
\end{proposition}

Setting $\tau=\eta/2$, Theorem~\ref{thm:screening} and Proposition~\ref{prop:active_connectivity} yield a direct local generation procedure. Starting from $\hat v$, we enumerate adjacent vertices, retain those satisfying $\hat c^\top(v-\hat v)\leq\eta/2,$ and recursively expand every newly retained vertex. Because the induced active subgraph is connected, exhaustive traversal recovers every active piece without expanding an inactive vertex. We call this procedure a \emph{margin-truncated adjacency search}.

Given an adjacency oracle, the search expands only active vertices while inspecting their incident edges. Its computational effort therefore scales with the explored active subgraph and the oracle cost, rather than with the full vertex set. A best-first search prioritizes vertices by $\Delta(v):=\hat c^\top(v-\hat v),$ thereby enumerating active vertices in nondecreasing predicted gap and enabling certified early termination.

This ordering also enables a certified early-stopping rule. For $0\leq\tau\leq\eta/2$, define the truncated active set and its restricted regret adversary as
\begin{equation}
\begin{aligned}
    \mathcal A_\tau
    &:=
    \{k\in\mathcal A:\Delta_k\leq\tau\},\\
    \rho_{\eta,\tau}(x)
    &:=
    \max_{k\in\mathcal A_\tau}\phi_k(x).
\end{aligned}
\label{eq:truncated_adversary}
\end{equation}

\begin{corollary}[Certified gap-ordered truncation]
\label{cor:gap-ordered-truncation}
For every $x\in\mathcal X$, the contribution of the omitted active pieces satisfies
\begin{equation*}
\max_{k\in\mathcal A\setminus\mathcal A_\tau}\phi_k(x)\leq B_\tau(x):=
\left(\eta+2\hat c^\top(x-\hat v)-2\tau\right)_+,
\end{equation*}
where the maximum over an empty index set is defined as zero.

Let $x_\tau\in\arg\min_{x\in\mathcal D}\rho_{\eta,\tau}(x), R_\tau:=\rho_{\eta,\tau}(x_\tau)$, and let $R_{\mathcal D}^\star:=\min_{x\in\mathcal D}\rho_\eta(x)$ denote the full refinement value over the same decision domain. Then
\begin{equation}
R_\tau \leq R_{\mathcal D}^\star \leq \rho_\eta(x_\tau) \leq \max\{R_\tau,B_\tau(x_\tau)\}.
\end{equation}
Consequently, $B_\tau(x_\tau)\leq R_\tau$ certifies that $x_\tau$ is optimal for the full refinement problem.
\end{corollary}

Corollary~\ref{cor:gap-ordered-truncation} turns best-first enumeration into a certified anytime procedure. At any threshold $\tau$, it provides a lower bound $R_\tau$ on the full optimum and an upper bound $\max\{R_\tau,B_\tau(x_\tau)\}$ on the full-set regret of the current truncated solution. The search may terminate with a certificate of exact optimality when $B_\tau(x_\tau)\leq R_\tau$, or with an approximate certificate when the resulting certified optimality gap falls below a prescribed tolerance.

\begin{figure*}[t]
    \centering
    \includegraphics[width=0.76\textwidth]{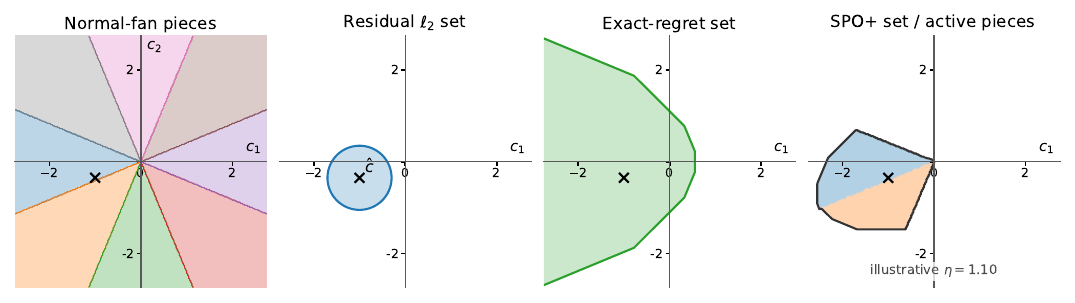}
    \caption{Geometry of residual and regret sets in cost coordinates $(c_1,c_2)$.  The panels show normal-fan regions, a residual-$\ell_2$ ball, an exact-regret set, and the active pieces of an SPO$+$ set.}
    \label{fig:section6_geometry}
\end{figure*}

\subsection{Certified Regret Refinement}
\label{sec:certified_refinement}

The connectivity result also yields a locally computable margin
certificate. Let $\operatorname{Adj}(\hat v)$ denote the vertices
adjacent to $\hat v$, and define
\begin{equation}
    \Gamma:
    =\min_{v\in\mathcal V\setminus\{\hat v\}}\hat c^\top(v-\hat v)
    =\min_{v\in\operatorname{Adj}(\hat v)}\hat c^\top(v-\hat v).
    \label{eq:local_predicted_margin}
\end{equation}
The second equality follows from Proposition~\ref{prop:active_connectivity}. Hence, $\Gamma$ is available from the initial adjacency expansion and measures the predicted separation from the closest competing vertex.

We consider two instances of \eqref{eq:generic-regret-refinement}: the pure refinement, restricted to vertices, and the continuous refinement over the full feasible polytope:
\begin{align}
    x^{\mathrm{pure}} &\in\arg\min_{x\in\mathcal V}\rho_\eta(x),
    & R_{\mathrm{pure}} &:=\rho_\eta(x^{\mathrm{pure}}),\label{eq:pure_refinement}\\
    x^{\mathrm{cont}} &\in\arg\min_{x\in\mathcal X}\rho_\eta(x),
    & R_{\mathrm{cont}}
    &:=\rho_\eta(x^{\mathrm{cont}}).
    \label{eq:continuous_refinement}
\end{align}

For $q\in\{\mathrm{pure, cont}\}$, let $x^q_\mathrm{test}$ and $R^q_\mathrm{test}$ denote the corresponding refined decision and robust regret value at $\hat{c}_\mathrm{test}$.
\begin{theorem}[Margin-certified regret refinement]
\label{thm:margin_refinement}
The two refinement problems attain their optima and satisfy
\begin{equation}
    0\leq R_{\mathrm{cont}}\leq R_{\mathrm{pure}}\leq\rho_\eta(\hat v)
    \leq(\eta-2\Gamma)_+\leq\eta,
    \label{eq:refinement_ordering}
\end{equation}
where $(a)_+:=\max\{a,0\}$. Moreover, setting $\eta=\eta_\alpha$, under the exchangeability condition of Theorem~\ref{thm:spo-plus-validity},
\begin{equation*}
\begin{aligned}
\mathbb P\!\Bigl\{&
    \operatorname{Reg}(x^q_{\mathrm{test}},c_{\mathrm{test}})
    \leq R^q_{\mathrm{test}}\\
&\leq (\eta_\alpha-2\Gamma_{\mathrm{test}})_+,
    \quad \forall q\in\{\mathrm{pure},\mathrm{cont}\}
\Bigr\}
\geq 1-\alpha.
\end{aligned}
\end{equation*}
\end{theorem}

The deterministic ordering shows that both refinements weakly improve the nominal worst-case certificate, while the larger continuous decision domain cannot perform worse than the vertex-restricted domain. 
The predicted margin further tightens the calibrated threshold from
$\eta$ to $(\eta-2\Gamma)_+$. In particular,
\begin{equation}
    \eta\leq2\Gamma
    \quad\Longrightarrow\quad
    \rho_\eta(\hat v)=0.
    \label{eq:zero_certificate_condition}
\end{equation}
Because $\Gamma$ is available from the initial adjacency expansion, this condition provides an exact early exit: the nominal decision is already certified and refinement is unnecessary. Otherwise, $(\eta-2\Gamma)_+$ quantifies the remaining worst-case certificate. With $\eta=\eta_\alpha$, the same bound controls the realized regret of
both refinements with probability at least $1-\alpha$.

\paragraph{Computational Formulations.}
Continuous refinement is natural for fractional or splittable decisions, whereas pure refinement preserves vertex-valued prescriptions. Since $\mathcal V\subseteq\mathcal X, R_\mathrm{cont}\le R_\mathrm{pure}$; however, this worst-case ordering need not extend to realized mean or tail regret. Moreover, $x^\mathrm{pure}$ need not be active, because active vertices index adversarial pieces rather than candidate decisions.

Suppose piece $P_k$ has constraint matrix $H_k$ and right-hand-side vector $h_k$:
\begin{equation}
    P_k=\{c\in\mathbb{R}^{n}:H_kc\leq h_k\}.
    \label{eq:piece_matrix_form}
\end{equation}
With nonnegative dual multipliers $y_k$, strong duality gives
\begin{equation}
    \phi_k(x)=\min_{y_k\geq0}
    \left\{h_k^\top y_k:H_k^\top y_k=x-v_k\right\}.
    \label{eq:piece_dual_lp}
\end{equation}
Aggregating these dual representations over the active pieces, with a scalar upper bound $t$ on worst-case regret, yields the following finite continuous master.
\begin{proposition}[Finite continuous master]
\label{prop:continuous_master}
The continuous refinement value $R_{\mathrm{cont}}$ equals the optimal value of the linear program
\begin{align}
    \min_{x\in\mathcal{X},\,t,\,\{y_k\}_{k\in\mathcal{A}}} \quad & t
    \label{eq:continuous_master}\\
    \text{\rm s.t.}\quad
        & H_k^\top y_k=x-v_k,
        && k\in\mathcal{A}, \nonumber\\
        & h_k^\top y_k\leq t,
        && k\in\mathcal{A}, \nonumber\\
        & y_k\geq0,
        && k\in\mathcal{A}. \nonumber
\end{align}
\end{proposition}

Thus, once the active pieces are explicitly represented, continuous refinement reduces to a single linear program whose size depends on the active set.

When the explicit piece descriptions are unavailable or prohibitively large, the original constraint representation of $\mathcal{X}=\{x\in\mathbb{R}^{n}_{+}:Ax=b\}$ yields a fixed-piece oracle that avoids all comparator-vertex constraints.

\begin{proposition}[Constraint-based fixed-piece oracle]
\label{prop:fixed_piece_oracle}
For any fixed $\bar x\in\mathcal{X}$ and active vertex $v_k$, $\phi_k(\bar x)$ equals the optimal value of
\begin{align}
    \max_{c,\mu,\lambda}\quad
        & c^\top(\bar x-v_k)
    \label{eq:fixed_piece_oracle}\\
    \text{\rm s.t.}\quad
        & A^\top\mu\leq c,
        \qquad
        b^\top\mu=c^\top v_k,
        \nonumber\\
        & A^\top\lambda\geq c-2\hat c,
        \nonumber\\
        & b^\top\lambda
        -(c-2\hat c)^\top v_k
        \leq\eta,
        \nonumber\\
        & \mu,\lambda\in\mathbb{R}^{m}.
        \nonumber
\end{align}
\end{proposition}

Here $\mu$ and $\lambda$ are dual variables for the hindsight minimization and the maximization in the SPO+ score, certifying optimality of $v_k$ under $c$ and the SPO+ threshold, respectively. At a master incumbent $(\bar x,\bar t)$, an optimizer $c_k^\star$ of \eqref{eq:fixed_piece_oracle} yields the valid cut $t\geq c_k^{\star\top}(x-v_k)$ whenever $c_k^{\star\top}(\bar x-v_k)>\bar t$.

Hence, Propositions~\ref{prop:continuous_master} and \ref{prop:fixed_piece_oracle} are complementary: explicit pieces yield a single LP, whereas the oracle supports constraint generation for either the continuous or vertex-restricted master directly from $(A,b)$. Combined with screening and adjacency generation, they avoid monolithic optimization over the nonconvex conformal set.

\section{Experiments}
\label{sec:experiments}
We evaluate the proposed framework using an illustrative example and contextual fractional-knapsack experiments, focusing on set geometry, finite-sample validity, decision performance, and certified active-piece generation.

\subsection{Illustrative Geometry of Calibrated Sets}
We consider a two-dimensional decision problem with feasible set $\mathcal V$ given by the eight vertices of a regular octagon and objective $\min_{v\in\mathcal{V}}c^\top v$ after $c$ is realized. In Figure~\ref{fig:section6_geometry}, each point $c=(c_1,c_2)$ denotes a possible ground-truth cost, while the black cross marks the prediction $\hat c=(-1,-0.35)$, which selects the nominal action $\hat v$.  From left to right, the panels show (i) the hindsight-optimal action for each $c$, partitioning cost space into normal-fan regions; (ii) the conventional residual-$\ell_2$ ball of radius $0.70$ centered at $\hat c$; (iii) the costs for which $\hat v$ incurs exact regret at most $1.10$; and (iv) the conformal regret set with SPO$+$ score at the same illustrative threshold, with colors identifying its active normal-fan pieces. The comparison illustrates both the piecewise-convex structure of the SPO+ set and why only predicted near-optimal vertices need to be generated.

\subsection{Experiments on the Fractional Knapsack Problem}

Following \citet{ho-nguyen2022risk}, we generate nonlinear context-dependent item values with multiplicative noise and use their negatives as minimization costs. Complete implementation details are provided in the appendix. Small instances permit exhaustive geometric validation, while larger instances evaluate decision performance and computational efficiency.

\subsubsection{Active Pieces and Margin Early Exit}

\begin{figure}[!t]
    \centering
    \includegraphics[width=0.76\linewidth]{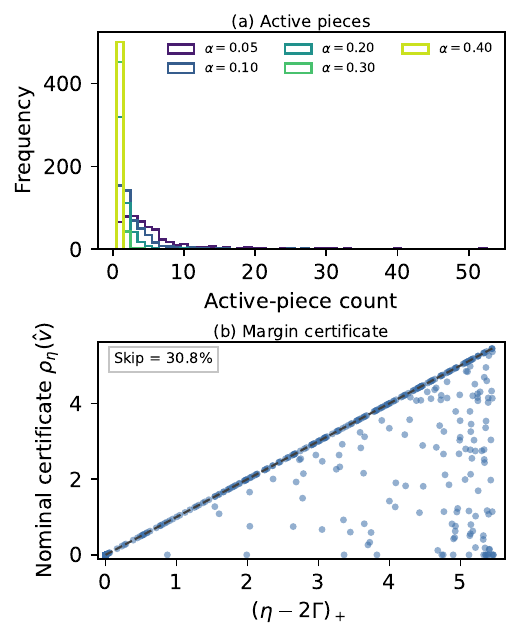}
    \caption{Exact geometry validation for $n=8$.  
    (a) Active-piece counts across calibration levels.
    (b) Nominal certificate versus the margin upper bound at $\alpha=0.10$; ``Skip'' denotes the fraction certified at zero without solving a robust master.}
    \label{fig:validation_active_margin}
\end{figure}

We first use $n=8$ instances, for which all vertices can be enumerated.
Although the feasible polytope contains $320.3$ vertices on average,
the mean number of active pieces ranges from only $5.33$ to $1.00$
across the calibration levels; see
Figure~\ref{fig:validation_active_margin}(a).

The adjacency procedure matches exhaustive enumeration for the active set and second-best margin in all context--level comparisons. At $\alpha=0.10$, all nominal certificates satisfy the margin bound within numerical tolerance, and $30.8\%$ of contexts receive a zero certificate (see Theorem~\ref{thm:margin_refinement}), allowing the robust master to be skipped. These results validate the exactness and local sparsity of the procedure.

\subsubsection{Coverage of Calibrated Sets.}

Parameter coverage measures inclusion of the realized cost in the calibrated set; decision-regret coverage measures whether realized regret is bounded by the set-based certificate. Table~\ref{tab:set_coverage_main} reports both metrics on the exact $n=8$ test split, with the decisions for the ``Regret set'' obtained using continuous regret refinement (RR). Detailed definitions are given in Appendix~\ref{app:set_coverage_comparison}.

\begin{table}[htb]
\centering
\footnotesize
\setlength{\tabcolsep}{3.2pt}
\renewcommand{\arraystretch}{0.96}
\caption{Empirical parameter / decision-regret coverage.}
\label{tab:set_coverage_main}
\begin{tabular}{@{}lccc@{}}
\toprule
Method & $\alpha=.05$ & $\alpha=.10$ & $\alpha=.20$ \\
\midrule
Regret set & 0.96/0.97 & 0.91/0.93 & 0.83/0.86 \\
Residual-$\ell_2$ set & 0.94/1.00 & 0.89/1.00 & 0.80/1.00 \\
Box set & 0.94/1.00 & 0.90/1.00 & 0.80/1.00 \\
KNN--Ellipsoid & 0.94/1.00 & 0.87/1.00 & 0.78/1.00 \\
\bottomrule
\end{tabular}
\end{table}

All four methods attain parameter coverage consistent with the nominal levels. The conventional sets nevertheless give essentially $100\%$ decision-regret coverage, which offers little information about decision risk because their full-set regret certificates are substantially wider: parameter inclusion implies decision-regret coverage, whereas the converse need not hold. By contrast, the regret set keeps the two coverage notions more closely aligned, providing a more decision-informative description of uncertainty rather than achieving high coverage through conservative bounds. Complete five-level results and certificate scales are reported in appendix.

\subsubsection{Decision Performance against Baselines}

We next evaluate decision performance on larger instances with $n=20$ and $\alpha=0.10$. We compare Nominal PTO with pure and continuous SPO$+$ regret refinement, as well as conformal Residual-$\ell_2$, conformal contextual Box, and KNN--Ellipsoid Cost-RO. Here we follow each method's intended objective: worst-case cost for RO and minimax regret for SPO+. Adapting the RO baselines to minimax regret goes beyond their standard formulations and can require more involved optimization. To isolate the role of the calibrated set, Appendix~\ref{app:objective_matched} reports objective-matched comparisons, which retain the performance advantage of SPO+ refinement.

\begin{table}[htb]
\centering
\footnotesize
\setlength{\tabcolsep}{2.6pt}
\renewcommand{\arraystretch}{0.96}
\caption{Decision performance at $\alpha=0.10$; lower regret is better. Q95 is the $0.95$ quantile, CVaR95 is the average regret in the upper $5\%$ tail, and Change is the fraction of decisions differing from nominal PTO.}
\label{tab:performance_main}
\begin{tabular}{@{}lrrrr@{}}
\toprule
Method & Mean & Q95 & CVaR95 & Change \\
\midrule
Nominal PTO & 7.605 & 38.664 & 60.557 & 0.000 \\
\textbf{Pure RR} & \textbf{7.763} & 38.956 & 61.814 & \textbf{0.100} \\
\textbf{Continuous RR} & 11.117 & \textbf{35.137} & \textbf{52.618} & 0.990 \\
Residual-$\ell_2$ RO & 20.001 & 57.090 & 77.027 & 0.956 \\
Box RO & 8.824 & 40.673 & 64.430 & 0.284 \\
KNN RO & 624.111 & 1374.573 & 1682.087 & 1.000 \\
\bottomrule
\end{tabular}
\end{table}

Continuous RR provides the strongest tail protection, reducing Q95 and $\operatorname{CVaR}_{0.95}$ by $9.1\%$ and $13.1\%$ relative to Nominal PTO, at the cost of higher mean regret. Pure RR changes only $10.0\%$ of decisions, remains close to the nominal mean, and achieves the best average performance among all robust methods. These results suggest using PTO when average performance is the priority, Pure RR for selective intervention with limited disruption to nominal decisions, and Continuous RR when upper-tail protection is worth a higher mean regret.
Additional experiments on real solar-grid data and sparse transportation LPs are reported in Appendix~\ref{app:additional_experiments}.

\begin{figure}[htb]
\centering
\includegraphics[width=0.76\linewidth]{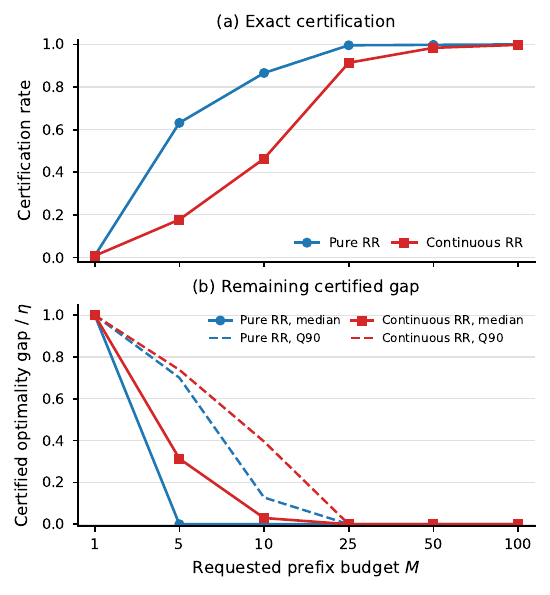}
\caption{Certified gap-ordered generation on $500$ test contexts.  (a) Exact-certification rate.  (b) Median and $0.90$ quantile of the certified optimality gap normalized by $\eta$.}
\label{fig:anytime_certification}
\end{figure}

\subsubsection{Certified Early Termination.}
We finally evaluate the practical effectiveness of the gap-ordered truncation certificate in Corollary~\ref{cor:gap-ordered-truncation} on the main $n=20$, $\alpha=0.10$ instances. On average, each feasible polytope has $1.16\times 10^6$ vertices, of which only $44.96$ generate active pieces. Let $M$ denote the number of generated active pieces.
As shown in Figure~\ref{fig:anytime_certification}, at $M=25$, exact optimality is already certified for $99.6\%$ of Pure RR and $91.4\%$ of Continuous RR instances, although only $56.2\%$ of active sets have been fully enumerated. The rates increase to $99.8\%$ and $98.4\%$, respectively, at $M=50$, while both the median and $0.90$ quantile of the normalized certified gap are already zero at $M=25$. Thus, the certificate is typically tight enough to establish global optimality well before complete active-set generation. Detailed budget-wise results and runtimes are reported in Appendix~\ref{app:anytime_piece_generation}.

\section{Conclusion}
\label{sec:conclusion}
We introduced Conformal Regret Sets, a decision-focused framework that calibrates SPO+ scores to control downstream regret rather than residuals. Split-conformal calibration provides a finite-sample marginal guarantee for the nominal PTO regret, while the induced nonconvex set admits a normal-fan-aligned polyhedral decomposition with exact active-piece screening, local generation, certified truncation, and margin-based certificates. These properties support certified pure and continuous regret-robust refinement without monolithic optimization. Fractional-knapsack experiments validate the theory: pure refinement preserves near-nominal average performance, continuous refinement improves upper-tail protection, and gap-ordered generation often certifies optimality before the full active set is recovered.

\bibliography{references}

\clearpage
\appendix
\section{Proofs and extended analysis}
\label{app:proofs_extended}

Unless stated otherwise, the geometric arguments below fix a prediction $\hat c$, a finite threshold $\eta\geq0$, and a nonempty compact polytope $\mathcal X=\operatorname{conv}(\mathcal V)$ with complete vertex set $\mathcal V=\{v_1,\ldots,v_K\}$.  We write $[K]:=\{1,\ldots,K\}$, let $\hat v:=x^*(\hat c)$, and let $\hat k\in[K]$ satisfy $v_{\hat k}=\hat v$.  We use $\operatorname{Adj}(v)$ for the vertices adjacent to $v$ in the one-skeleton of $\mathcal X$.  In proofs concerning \eqref{eq:poly_decompose}, $\eta$ denotes the generic threshold fixed in the main paper's geometry section; it is specialized to $\eta_\alpha$ for the calibrated set.  All dimensions below follow the main paper: costs and decisions lie in $\mathbb R^n$. Throughout the Main Paper subsection, the main text's unique-optimizer convention is in force, and we abbreviate $\rho_\eta(x;\hat c)$ as $\rho_\eta(x)$.  Full optimizer-tie statements are deferred to the Extended Analysis subsection; tie-boundary closures are used below only to justify
the exact worst-case values.

\subsection{Main Paper}

\paragraph{Exact-SPO degeneracy and SPO+ dominance.}

\begin{proof}[Proof of Proposition~\ref{prop:spo_robustification_degeneracy}]
The nominal bound is immediate from the definition of the exact-SPO
sublevel set.  For every
$c\in\mathcal U_{\mathrm{SPO}}(\hat c;\eta)$,
\[
 \operatorname{Reg}(\hat x,c)
 =c^\top\hat x-\min_{u\in\mathcal X}c^\top u
 =s_{\mathrm{SPO}}(c,\hat c)
 \leq\eta.
\]
Taking the supremum proves the nominal bound in the proposition.

It remains to show that every alternative decision has unbounded regret.
Because $\hat x$ is a vertex of the polyhedron $\mathcal X$, it is an
exposed point: there exists $d\in\mathbb R^n$ such that
\begin{equation}
 d^\top(x-\hat x)>0
 \qquad\text{for every }x\in\mathcal X\setminus\{\hat x\}.
 \label{eq:exposing_direction_exact_spo}
\end{equation}
For completeness, choose $d$ in the relative interior of the minimization
normal cone of $\hat x$.  The face--normal-cone correspondence for
polyhedra states that the minimizers exposed by any vector in this relative
interior form exactly the face associated with that normal cone.  Since the
normal cone is the one associated with the vertex $\hat x$, the exposed
face is the singleton $\{\hat x\}$, which proves
\eqref{eq:exposing_direction_exact_spo}.

For $t>0$, set $c_t:=td$.  Equation
\eqref{eq:exposing_direction_exact_spo} makes $\hat x$ the unique optimizer
under $c_t$, and hence
\[
 s_{\mathrm{SPO}}(c_t,\hat c)
 =c_t^\top\hat x-\min_{u\in\mathcal X}c_t^\top u=0.
\]
Thus $c_t\in\mathcal U_{\mathrm{SPO}}(\hat c;\eta)$ for every $t>0$.
For any fixed $x\neq\hat x$,
\[
 \operatorname{Reg}(x,c_t)
 =t\,d^\top(x-\hat x)\longrightarrow+\infty.
\]
This proves the asserted unboundedness.  The robust value at
$\hat x$ is finite, whereas it is infinite at every other feasible
decision, so $\hat x$ is the unique robust minimizer.
\end{proof}

We next record explicitly the pointwise inequality used by the conformal
argument.  Let $v:=x^*(c)$ and $\hat v:=x^*(\hat c)$.  Their optimality gives
\begin{equation}
 c^\top(\hat v-v)\geq0,
 \qquad
 \hat c^\top(\hat v-v)\leq0.
 \label{eq:two_optimality_signs_new}
\end{equation}
Selecting $x=v$ in the maximum defining $s_{\mathrm{SPO+}}$ gives
$s_{\mathrm{SPO+}}(c,\hat c)\geq0$.  Selecting $x=\hat v$ instead gives
\begin{align}
 s_{\mathrm{SPO+}}(c,\hat c)
 &\geq(c-2\hat c)^\top(\hat v-v) \nonumber\\
 &=\operatorname{Reg}(\hat v,c)
   -2\hat c^\top(\hat v-v) \nonumber\\
 &\geq\operatorname{Reg}(\hat v,c)
 =s_{\mathrm{SPO}}(c,\hat c).
 \label{eq:spop_dominates_spo_new}
\end{align}
The last inequality has the stated direction because the second inner
product in~\eqref{eq:two_optimality_signs_new} is nonpositive.

\begin{proof}[Proof of Theorem~\ref{thm:spo-plus-validity}]
We use the standard split-sample interpretation of the theorem: the
training data, and hence the fitted predictor, are independent of the joint
calibration--test tuple.  Condition on those training data.  The fitted
predictor and all deterministic measurable optimization maps are then
fixed functions, while the calibration and test observations remain
exchangeable.  Applying the same SPO+ score map to them therefore produces
exchangeable scores $S_1,\ldots,S_N,S_{N+1}$.  Independence from the
calibration sample alone would not suffice if the fitted predictor were
allowed to depend on the test observation.

First suppose these $N+1$ scores are distinct and
$q_\alpha\leq N$.  The rank $R$ of $S_{N+1}$ among all $N+1$ scores is
uniform on $\{1,\ldots,N+1\}$.  Consequently,
\[
 \mathbb P\{R\leq q_\alpha\}
 =\frac{q_\alpha}{N+1}\geq1-\alpha.
\]
If $R\leq q_\alpha$, fewer than $q_\alpha$ calibration scores are smaller
than $S_{N+1}$, so
$S_{N+1}\leq S_{(q_\alpha)}=\eta_\alpha$.  Hence
\[
 \mathbb P\{S_{N+1}\leq\eta_\alpha\}\geq1-\alpha,
\]
which is exactly~\eqref{eq:spo-plus-set-validity}.  When
$q_\alpha=N+1$, the convention $\eta_\alpha=+\infty$ makes the inclusion
event certain.  This last branch is a valid but vacuous statistical
statement: all recession, screening, and finite-refinement results below
require $\eta_\alpha<+\infty$.  Calibration-score ties are handled without
a continuity assumption in the extended analysis below.

On the same score-inclusion event,
\eqref{eq:spop_dominates_spo_new} yields
\[
 \operatorname{Reg}\!\left(
 x^*(\hat c_{\mathrm{test}}),c_{\mathrm{test}}\right)
 \leq s_{\mathrm{SPO+}}\!\left(
 c_{\mathrm{test}},\hat c_{\mathrm{test}}\right)
 \leq\eta_\alpha.
\]
The event on the left therefore has probability at least $1-\alpha$,
proving the second assertion.
\end{proof}

\paragraph{Polyhedral decomposition under the main-paper convention.}

\begin{proof}[Proof of Theorem~\ref{thm:decomposition}]
Consider first a cost $c$ having the unique optimizer $v_k$.  Since
$\mathcal X=\operatorname{conv}(\mathcal V)$, a linear objective attains
the same maximum over $\mathcal X$ and over $\mathcal V$.  Thus
\begin{align}
 s_{\mathrm{SPO+}}(c,\hat c)
 &=\max_{1\leq j\leq K}
   (c-2\hat c)^\top(v_j-v_k).
 \label{eq:spop_vertex_max_decomposition}
\end{align}
The score is at most $\eta$ if and only if, for every $j$,
\[
 c^\top(v_j-v_k)
 \leq\eta+2\hat c^\top(v_j-v_k).
\]
Meanwhile, optimality of $v_k$ under the minimization objective is
equivalent to
\[
 c^\top(v_k-v_j)\leq0
 \qquad\text{for every }j.
\]
These are precisely the two families of inequalities defining $P_k$ in
\eqref{eq:poly_decompose}.  Hence, at every cost with a unique optimizer,
\[
 c\in\mathcal U_{\mathrm{SPO+}}(\hat c;\eta)
 \quad\Longleftrightarrow\quad
 c\in\bigcup_{k=1}^K P_k.
\]
This proves the pointwise tie-free decomposition.

We next justify the use of the closed pieces in the subsequent worst-case
regret values.  Let $\mathcal U_\eta^{\mathrm{uni}}$ denote the subset of
$\mathcal U_{\mathrm{SPO+}}(\hat c;\eta)$ consisting of costs whose linear
objective has a unique minimizer over $\mathcal X$.
Under the main paper's unique-nominal-optimizer assumption,
\begin{equation}
 \operatorname{cl}\!\left(\mathcal U_\eta^{\mathrm{uni}}\right)
 =\bigcup_{k=1}^K P_k.
 \label{eq:unique_region_closure_new}
\end{equation}
The inclusion from left to right follows from the pointwise equivalence
above and the fact that a finite union of closed polyhedra is closed.  For
the reverse inclusion, fix a nonempty $P_k$ and write
$\Delta_j:=\hat c^\top(v_j-\hat v)$.  Using $v_j=\hat v$ in the two
families defining $P_k$ gives
\[
 0\leq c^\top(\hat v-v_k)\leq\eta-2\Delta_k
 \qquad(c\in P_k),
\]
so $\Delta_k\leq\eta/2$.

Suppose first that $\eta>0$ and $\Delta_k<\eta/2$.  Choose an exposing
cost $d_k$ for which $v_k$ is the unique minimizer, so
$d_k^\top(v_j-v_k)>0$ for every $j\neq k$.  The right-hand side of each
upper piece inequality satisfies
\[
 \eta+2\hat c^\top(v_j-v_k)
 =\eta+2(\Delta_j-\Delta_k)
 \geq\eta-2\Delta_k>0.
\]
Consequently, for all sufficiently small $\varepsilon>0$,
$\varepsilon d_k\in P_k$, and all its lower piece inequalities are strict
for $j\neq k$.  For any $c\in P_k$ and $t\in(0,1]$, convexity of $P_k$
then gives
\[
 c_t:=(1-t)c+t\varepsilon d_k\in P_k,
\]
with $v_k$ uniquely optimal under $c_t$.  The pointwise equivalence already
proved implies $c_t\in\mathcal U_\eta^{\mathrm{uni}}$, and $c_t\to c$ as
$t\downarrow0$.

If $\eta>0$ and $\Delta_k=\eta/2$, the displayed two-sided bound forces
$c^\top(\hat v-v_k)=0$ throughout $P_k$.  Thus $\hat v$ is also optimal
under every $c\in P_k$.  Moreover, for every $j$,
\begin{align*}
 c^\top(v_j-\hat v)
 &=c^\top(v_j-v_k)\\
 &\leq\eta+2\hat c^\top(v_j-v_k)
 =2\hat c^\top(v_j-\hat v)\\
 &\leq\eta+2\hat c^\top(v_j-\hat v),
\end{align*}
so $P_k\subseteq P_{\hat k}$.  Since
$\Delta_{\hat k}=0<\eta/2$, the preceding argument applies to
$P_{\hat k}$.

Finally, if $\eta=0$, nonemptiness forces $\Delta_k=0$.  Uniqueness of
$\hat v$ under $\hat c$ then implies $k=\hat k$.  The cost $\hat c$ itself
belongs to $P_{\hat k}$ and uniquely exposes $\hat v$, so mixing any
$c\in P_{\hat k}$ with a positive amount of $\hat c$ again produces points
of $\mathcal U_\eta^{\mathrm{uni}}$ converging to $c$.  This proves
\eqref{eq:unique_region_closure_new}.

For any deterministic vertex-valued extension of $x^*(c)$ to tie costs, the same
inequality calculation used above gives
\[
 \mathcal U_\eta^{\mathrm{uni}}
 \subseteq\mathcal U_{\mathrm{SPO+}}(\hat c;\eta)
 \subseteq\bigcup_{k=1}^K P_k.
\]
Thus all three sets have the same closure.  The regret
$c^\top x-\min_{u\in\mathcal X}c^\top u$ is continuous in $c$, so taking
its supremum over any of these sets gives the same value.  Finally, every
$c\in P_k$ makes $v_k$ optimal and hence satisfies
$\operatorname{Reg}(x,c)=c^\top(x-v_k)$.  It follows that
\[
 \rho_\eta(x)=\max_{k\in\mathcal A}\phi_k(x),
\]
which therefore justifies
\eqref{eq:piecewise_regret_representation}.  The unclosed sets themselves
can differ on optimizer-tie boundaries; the precise extension is given in
Proposition~\ref{prop:tie_safe_extension}.
\end{proof}

\paragraph{Recession directions, screening, and active-piece generation.}

\begin{proof}[Proof of Proposition~\ref{prop:finite_piece_regret}]
Fix a nonempty piece $P_k$ and abbreviate
\[
 a_j:=v_j-v_k,
 \qquad
 b_j:=\eta+2\hat c^\top(v_j-v_k).
\]
The two families of inequalities in~\eqref{eq:poly_decompose} can be
written as
\begin{equation}
 P_k=\{c:-a_j^\top c\leq0,
                 a_j^\top c\leq b_j,\ j=1,\ldots,K\}.
 \label{eq:piece_paired_rows_new}
\end{equation}
Because $P_k$ is nonempty, $d$ is a recession direction if and only if it
satisfies the homogeneous version of every row in
\eqref{eq:piece_paired_rows_new}.  Thus
\[
 -a_j^\top d\leq0
 \quad\text{and}\quad
 a_j^\top d\leq0
 \qquad\text{for every }j,
\]
which is equivalent to
\begin{equation}
 d^\top(v_j-v_k)=0
 \qquad\text{for every }j.
 \label{eq:recession_vertex_orthogonality_new}
\end{equation}

We next verify the required span identity.  The inclusion
\[
 \operatorname{span}\{v_j-v_k:j\in[K]\}
 \subseteq\operatorname{span}(\mathcal X-\mathcal X)
\]
is immediate.  Conversely, write any $x,y\in\mathcal X$ as
$x=\sum_j\lambda_jv_j$ and $y=\sum_j\mu_jv_j$, where both coefficient
vectors are nonnegative and sum to one.  Then
\[
 x-y=\sum_j(\lambda_j-\mu_j)(v_j-v_k),
\]
because $\sum_j(\lambda_j-\mu_j)=0$.  Hence
\[
 \operatorname{span}\{v_j-v_k:j\in[K]\}
 =\operatorname{span}(\mathcal X-\mathcal X).
\]
Combining this identity with
\eqref{eq:recession_vertex_orthogonality_new} proves
\eqref{eq:piece_recession}.

Now fix $x\in\mathcal X$ and set $q:=x-v_k$.  Then
$q\in\operatorname{span}(\mathcal X-\mathcal X)$, so
$q^\top d=0$ for every $d\in\operatorname{rec}(P_k)$.  To establish
finiteness and attainment rather than only the absence of an improving
ray, take a finite-generator representation
\[
 P_k=\operatorname{conv}\{p_1,\ldots,p_r\}
     +\operatorname{cone}\{r_1,\ldots,r_s\}+L_k,
\]
where $r\geq1$, every $r_\ell$ is a recession direction, and the
lineality space $L_k$ is contained in $\operatorname{rec}(P_k)$.  For
arbitrary $c\in P_k$,
\[
 c=\sum_i\theta_ip_i+\sum_\ell\gamma_\ell r_\ell+\ell.
\]
Here $\theta_i\geq0$, $\sum_i\theta_i=1$,
$\gamma_\ell\geq0$, and $\ell\in L_k$.
Orthogonality to the recession cone yields
\[
 q^\top c=\sum_i\theta_iq^\top p_i
 \leq\max_{1\leq i\leq r}q^\top p_i.
\]
The upper bound is finite and is attained by a maximizing generator
$p_i\in P_k$.  This proves that $\phi_k(x)$ in
\eqref{eq:piece_regret} is finite and attained.  The same representation
also gives
\begin{equation}
 \phi_k(x)=\max_{1\leq i\leq r}p_i^\top(x-v_k),
 \label{eq:piece_support_finite_max_new}
\end{equation}
so $\phi_k$ is a continuous polyhedral convex function on $\mathcal X$.
\end{proof}

\begin{proof}[Proof of Theorem~\ref{thm:screening}]
Suppose first that $P_k$ is nonempty and choose $c\in P_k$.  Use
$v_j=\hat v$ in the lower and upper inequalities defining the piece.  Since
$\Delta_k=\hat c^\top(v_k-\hat v)$,
\begin{align}
 0
 &\leq c^\top(\hat v-v_k) \nonumber\\
 &\leq\eta+2\hat c^\top(\hat v-v_k)
 =\eta-2\Delta_k.
 \label{eq:screening_necessity_chain_new}
\end{align}
The two endpoints imply $\Delta_k\leq\eta/2$.

Conversely, suppose $\Delta_k\leq\eta/2$ and set $c=0$.  Nominal
optimality of $\hat v$ gives
$\hat c^\top(v_j-\hat v)\geq0$ for every $j$.  Therefore
\begin{align}
 \eta+2\hat c^\top(v_j-v_k)
 &=\eta+2\hat c^\top(v_j-\hat v)-2\Delta_k \nonumber\\
 &\geq\eta-2\Delta_k\geq0.
 \label{eq:screening_zero_witness_new}
\end{align}
At $c=0$, both $c^\top(v_j-v_k)$ and
$c^\top(v_k-v_j)$ are zero, and
\eqref{eq:screening_zero_witness_new} verifies every remaining upper
bound.  Hence $0\in P_k$, proving the reverse implication and
\eqref{eq:exact_active_screen}.  In particular,
$P_{\hat k}$ is always nonempty because $\Delta_{\hat k}=0$.
This argument concerns the closed algebraic pieces $P_k$.  At
$\Delta_k=\eta/2$, the zero witness can lie entirely on an optimizer-tie
boundary; Proposition~\ref{prop:tie_safe_extension} explains when these
pieces are exact score pieces and when they form a conservative outer
representation.
\end{proof}

\begin{proof}[Proof of Proposition~\ref{prop:active_connectivity}]
At a vertex $v$, the tangent cone of a polytope is generated by the
directions of its incident edges:
\[
 T_{\mathcal X}(v)
 =\operatorname{cone}\{u-v:u\in\operatorname{Adj}(v)\}.
\]
Suppose $v\neq\hat v$ and every adjacent vertex $u$ satisfied
$\hat c^\top(u-v)\geq0$.  Every vector $x-v$, $x\in\mathcal X$, belongs
to $T_{\mathcal X}(v)$, so the displayed conic representation would imply
$\hat c^\top(x-v)\geq0$ for every $x\in\mathcal X$.  Then $v$ would be a
global minimizer of $\hat c^\top x$, contradicting uniqueness of
$\hat v$.  Thus every $v\neq\hat v$ has an adjacent vertex with strictly
smaller predicted objective.

Starting at any vertex $v$, repeatedly take such a decreasing edge.
Because $\mathcal V$ is finite and the objective decreases strictly at
each step, the path terminates; its terminal vertex has no decreasing
neighbor and therefore must be the unique minimizer $\hat v$.  Reversing
the path gives
\[
 \hat v=v^0,v^1,\ldots,v^L=v,
 \qquad
 \hat c^\top v^0\leq\cdots\leq\hat c^\top v^L.
\]
If $v\in\mathcal V_\tau$, every preceding vertex on this path has
predicted objective no larger than that of $v$, and hence also belongs to
$\mathcal V_\tau$.  The induced subgraph is therefore connected, proving
\eqref{eq:monotone_vertex_path}.

The same path proves the local-margin identity
\eqref{eq:local_predicted_margin} when $|\mathcal V|\geq2$.  Indeed, because
$\operatorname{Adj}(\hat v)\subseteq
\mathcal V\setminus\{\hat v\}$,
\[
 \Gamma\leq
 \min_{u\in\operatorname{Adj}(\hat v)}
 \hat c^\top(u-\hat v).
\]
Conversely, the first step $v^1$ on a monotone path from $\hat v$ to any
$v\neq\hat v$ is adjacent to $\hat v$ and satisfies
$\hat c^\top(v^1-\hat v)\leq\hat c^\top(v-\hat v)$.  Minimizing over
$v\neq\hat v$ gives the reverse inequality.
\end{proof}

\paragraph{Certified truncation.}

\begin{proof}[Proof of Corollary~\ref{cor:gap-ordered-truncation}]
For the minimum notation in the statement to be literal, the decision
domain $\mathcal D$ is assumed nonempty and compact; this includes both
$\mathcal V$ and $\mathcal X$.  More generally, the displayed argument is
conditional on the stated minimizers existing.
Fix $x\in\mathcal X$ and write
$x=\sum_j\lambda_jv_j$, with $\lambda_j\geq0$ and
$\sum_j\lambda_j=1$.  For $c\in P_k$, summing the upper piece
inequalities with these weights gives
\begin{align}
 c^\top(x-v_k)
 &\leq\eta+2\hat c^\top(x-v_k) \nonumber\\
 &=\eta+2\hat c^\top(x-\hat v)-2\Delta_k.
 \label{eq:arbitrary_x_piece_bound_new}
\end{align}
The lower piece inequalities make $v_k$ optimal under $c$, so the
left-hand side is nonnegative.  Taking the maximum over $P_k$ and then taking the 
positive part in~\eqref{eq:arbitrary_x_piece_bound_new} yields
\[
 \phi_k(x)\leq
 \left(\eta+2\hat c^\top(x-\hat v)-2\Delta_k\right)_+.
\]
Every omitted active piece has $\Delta_k>\tau$.  Monotonicity of the
positive-part map therefore gives
\[
 \max_{k\in\mathcal A\setminus\mathcal A_\tau}\phi_k(x)
 \leq
 \left(\eta+2\hat c^\top(x-\hat v)-2\tau\right)_+
 =B_\tau(x).
\]
If the omitted set is empty, its contribution is defined as zero, and the
same inequality holds.

Because $\mathcal A_\tau\subseteq\mathcal A$,
$\rho_{\eta,\tau}(x)\leq\rho_\eta(x)$ for every $x$.
Consequently,
\[
 R_\tau
 \leq R_{\mathcal D}^\star
 \leq\rho_\eta(x_\tau).
\]
At $x_\tau$, the retained pieces contribute
$\rho_{\eta,\tau}(x_\tau)=R_\tau$, while all omitted pieces are bounded by
$B_\tau(x_\tau)$.  Hence
\[
 \rho_\eta(x_\tau)
 \leq\max\{R_\tau,B_\tau(x_\tau)\},
\]
which proves the full chain.  If $B_\tau(x_\tau)\leq R_\tau$, then
\[
 R_\tau
 \leq R_{\mathcal D}^\star
 \leq\rho_\eta(x_\tau)
 \leq R_\tau.
\]
Thus $x_\tau$ is optimal for the full problem.  Since
$\mathcal A_\tau$ uses the inclusive condition $\Delta_k\leq\tau$, a
certified gap-ordered implementation must exhaust the entire gap plateau at
the cutoff; stopping in the middle of a tied plateau does not produce the
set $\mathcal A_\tau$ used in this proof.
This identification with $\rho_\eta$ uses the main-text unique-optimizer
convention; Proposition~\ref{prop:tie_safe_extension} gives the corresponding
tie-safe statement.
\end{proof}

\paragraph{Margin certification and finite formulations.}

\begin{proof}[Proof of Theorem~\ref{thm:margin_refinement}]
For every active $k$ and every $c\in P_k$, the lower piece inequalities
make $v_k$ optimal under $c$.  Therefore, for all $x\in\mathcal X$,
\[
 c^\top(x-v_k)=\operatorname{Reg}(x,c)\geq0.
\]
Taking the attained maximum over $P_k$ and then the finite maximum over
$k\in\mathcal A$ shows that $\rho_\eta(x)\geq0$.

Equation~\eqref{eq:piece_support_finite_max_new} shows that every
$\phi_k$ is continuous.  Hence
$\rho_\eta=\max_{k\in\mathcal A}\phi_k$ is continuous as a finite
maximum.
The pure domain $\mathcal V$ is finite and the continuous domain
$\mathcal X$ is compact, so both refinement problems attain their minima.
Since
$\hat v\in\mathcal V\subseteq\mathcal X$, minimizing the same function
over these nested domains gives
\begin{equation}
 0\leq R_{\mathrm{cont}}
 \leq R_{\mathrm{pure}}
 \leq\rho_\eta(\hat v).
 \label{eq:refinement_first_chain_new}
\end{equation}

We now bound the last term piece by piece.  The nominal piece contributes
\[
 \phi_{\hat k}(\hat v)
 =\max_{c\in P_{\hat k}}c^\top(\hat v-\hat v)=0.
\]
For a nonnominal active vertex $v_k$, use $v_j=\hat v$ in the upper piece
inequality.  For every $c\in P_k$,
\begin{align}
 0
 &\leq c^\top(\hat v-v_k)
 =\operatorname{Reg}(\hat v,c) \nonumber\\
 &\leq\eta+2\hat c^\top(\hat v-v_k)
 =\eta-2\Delta_k
 \leq\eta-2\Gamma.
 \label{eq:nonnominal_margin_chain_new}
\end{align}
The last comparison is used only for $v_k\neq\hat v$; it is not valid for
the nominal index.  Taking the maximum of the zero nominal contribution and
the nonnominal bounds proves
\[
 \rho_\eta(\hat v)\leq(\eta-2\Gamma)_+\leq\eta.
\]
Together with~\eqref{eq:refinement_first_chain_new}, this proves
\eqref{eq:refinement_ordering}.  The singleton convention needed when
$|\mathcal V|=1$ is given in the extended analysis.

The equality boundary is included.  If $\eta=2\Gamma<+\infty$, every
active nonnominal vertex satisfies
$\Delta_k\geq\Gamma$ and, by Theorem~\ref{thm:screening},
$\Delta_k\leq\eta/2=\Gamma$.  Hence equality holds throughout and
\eqref{eq:nonnominal_margin_chain_new} forces every nonnominal contribution
at $\hat v$ to be zero.  The nominal contribution is also zero, so
$\rho_\eta(\hat v)=0$ and all terms preceding it in
\eqref{eq:refinement_ordering} are zero.

Finally set $\eta=\eta_\alpha<+\infty$ and consider the conformal inclusion
event from Theorem~\ref{thm:spo-plus-validity}.  On this event,
$c_{\mathrm{test}}\in\mathcal U_{\mathrm{SPO+}}
(\hat c_{\mathrm{test}};\eta_\alpha)$.
Therefore, for $q\in\{\mathrm{pure},\mathrm{cont}\}$,
\begin{align*}
 \operatorname{Reg}(x^q_{\mathrm{test}},c_{\mathrm{test}})
 &\leq\rho_{\eta_\alpha}(x^q_{\mathrm{test}})
 =R^q_{\mathrm{test}}\\
 &\leq(\eta_\alpha-2\Gamma_{\mathrm{test}})_+.
\end{align*}
Both choices of $q$ are controlled on the same inclusion event, whose
probability is at least $1-\alpha$.
\end{proof}

\begin{proof}[Proof of Proposition~\ref{prop:continuous_master}]
Fix an active piece $P_k=\{c:H_kc\leq h_k\}$ and a decision
$x\in\mathcal X$.  The support problem defining $\phi_k(x)$ is
\[
 \max_c\{(x-v_k)^\top c:H_kc\leq h_k\}.
\]
It is feasible because $P_k$ is active, and it has a finite optimum attained
 by Proposition~\ref{prop:finite_piece_regret}.  Its LP dual is
\[
 \min_{y_k\geq0}
 \{h_k^\top y_k:H_k^\top y_k=x-v_k\}.
\]
The equality has the displayed sign because the primal objective
coefficient of $c$ is $x-v_k$.  Strong duality gives equality of the two
attained optimal values and proves~\eqref{eq:piece_dual_lp}.

For fixed $x$ and $t$, weak duality shows that the existence of
$y_k\geq0$ satisfying
\[
 H_k^\top y_k=x-v_k,\qquad h_k^\top y_k\leq t
\]
implies $\phi_k(x)\leq t$.  Conversely, if $\phi_k(x)\leq t$, an optimal
dual solution exists and satisfies these two relations.  Imposing this
equivalence for every $k\in\mathcal A$ gives exactly the epigraph
\[
 t\geq\max_{k\in\mathcal A}\phi_k(x)=\rho_\eta(x)
\]
over $x\in\mathcal X$.  Minimizing $t$ therefore yields
$R_{\mathrm{cont}}$, proving~\eqref{eq:continuous_master}.
\end{proof}

\begin{proof}[Proof of Proposition~\ref{prop:fixed_piece_oracle}]
Fix $\bar x\in\mathcal X$ and $v_k\in\mathcal V$.  For a candidate cost
$c$, consider first the standard-form minimization problem
\[
 \min_{z\geq0}\{c^\top z:Az=b\}.
\]
Its dual is
\[
 \max_{\mu\in\mathbb R^m}
 \{b^\top\mu:A^\top\mu\leq c\}.
\]
Because $v_k$ is primal feasible, weak duality implies that the existence
of $\mu$ satisfying
\[
 A^\top\mu\leq c,\qquad b^\top\mu=c^\top v_k
\]
certifies optimality of $v_k$ under $c$.  Conversely, if $v_k$ is
optimal, LP strong duality supplies such a dual-optimal $\mu$.

Next set $d:=c-2\hat c$.  Conditional on $v_k$ being hindsight-optimal,
the SPO+ threshold is
\begin{equation}
 \max_{z\in\mathcal X}d^\top(z-v_k)\leq\eta.
 \label{eq:fixed_piece_threshold_primal_new}
\end{equation}
The dual of $\max\{d^\top z:Az=b,\ z\geq0\}$ is
\[
 \min_{\lambda\in\mathbb R^m}
 \{b^\top\lambda:A^\top\lambda\geq d\}.
\]
Nonemptiness and compactness of $\mathcal X$ make the primal optimum finite
and attained, so strong duality shows that
\eqref{eq:fixed_piece_threshold_primal_new} holds if and only if there is
$\lambda\in\mathbb R^m$ such that
\[
 A^\top\lambda\geq c-2\hat c,\qquad
 b^\top\lambda-(c-2\hat c)^\top v_k\leq\eta.
\]

The $\mu$ constraints therefore describe exactly the costs for which
$v_k$ is optimal, while the $\lambda$ constraints describe exactly the
SPO+ threshold for that same vertex.  Their projection onto the
$c$-coordinates is precisely $P_k$.  The objective in
\eqref{eq:fixed_piece_oracle} is
$c^\top(\bar x-v_k)$, so maximizing over the extended formulation has the
same value as maximizing over $c\in P_k$, namely $\phi_k(\bar x)$.  An
optimal $c$ exists by Proposition~\ref{prop:finite_piece_regret}, and
strong duality supplies corresponding $\mu$ and $\lambda$, so the oracle
also attains this value.
\end{proof}

\subsection{Extended Analysis}

\paragraph{Calibration-score ties and the augmented threshold.}
The split-conformal guarantee does not require a continuous score
distribution.  Conditional on training data that are independent of the
joint calibration--test tuple, let
$S_1,\ldots,S_N,S_{N+1}$ be the exchangeable scores.  Introduce independent
$U_1,\ldots,U_{N+1}\sim\operatorname{Unif}(0,1)$ and rank
$(S_i,U_i)$ lexicographically.  These pairs are exchangeable and almost
surely distinct, so the rank $R$ of the test pair is uniform on
$\{1,\ldots,N+1\}$.  For $q_\alpha\leq N$,
\[
 \mathbb P\{R\leq q_\alpha\}
 =\frac{q_\alpha}{N+1}\geq1-\alpha.
\]
If $S_{N+1}>S_{(q_\alpha)}$, at least $q_\alpha$ calibration scores are
strictly smaller than $S_{N+1}$.  Their pairs all precede the test pair,
so $R>q_\alpha$.  Taking the contrapositive gives
\[
 \{R\leq q_\alpha\}
 \subseteq\{S_{N+1}\leq S_{(q_\alpha)}\}.
\]
Thus the inclusive conformal event has probability at least
$1-\alpha$, including when calibration scores are tied.

If $q_\alpha=N+1$, the augmented threshold is $+\infty$ and coverage is
one.  This is only a vacuous statistical branch.  Even when
$\mathcal X$ is compact and every SPO+ score is finite, its
$+\infty$-sublevel set is all of $\mathbb R^n$; the upper piece
inequalities no longer restrict recession directions, and worst-case
regret can be infinite.  Accordingly, all screening, margin, and finite
master results are invoked only when $\eta_\alpha<+\infty$.

\paragraph{Optimizer ties and the closed piecewise adversary.}

\begin{proposition}[Tie-safe SPO+ representation and certificates]
\label{prop:tie_safe_extension}
Retain the main paper's unique nominal optimizer $\hat v$, but allow
realized costs to have optimizer ties and fix any deterministic measurable
vertex-valued hindsight selector.  For this extension only, define
\[
 \overline{\mathcal U}_\eta:=\bigcup_{k=1}^K P_k,
 \qquad
 \overline\rho_\eta(x):=
 \max_{k\in\mathcal A}\max_{c\in P_k}c^\top(x-v_k).
\]
For finite $\eta$,
\begin{equation}
 \mathcal U_{\mathrm{SPO+}}(\hat c;\eta)
 \subseteq\overline{\mathcal U}_\eta,
 \label{eq:tie_safe_set_containment_new}
\end{equation}
the two sets agree at every cost having a unique optimizer, and
\begin{equation}
 \operatorname{cl}\!\left(
 \mathcal U_{\mathrm{SPO+}}(\hat c;\eta)
 \right)
 =\overline{\mathcal U}_\eta.
 \label{eq:tie_safe_closure_new}
\end{equation}
Moreover, for every $x\in\mathcal X$,
\begin{equation}
 0\leq\rho_\eta(x)=\overline\rho_\eta(x)<+\infty.
 \label{eq:tie_safe_rho_order_new}
\end{equation}
Thus passing to the closed union does not change any worst-case regret
value, although it can add boundary costs and ensure that the piecewise adversarial maximum is
attained.  In particular, every additional cost in
$\overline{\mathcal U}_\eta\setminus
\mathcal U_{\mathrm{SPO+}}(\hat c;\eta)$ lies on an optimizer-tie boundary.

Define instead a compatible selector as follows: among all vertices
minimizing $c^\top v$, first minimize $\hat c^\top v$, and break any
remaining tie by a fixed deterministic vertex order.  If the SPO+ score
and its calibration scores use this same rule, then
\begin{equation}
 \mathcal U_{\mathrm{SPO+}}(\hat c;\eta)
 =\overline{\mathcal U}_\eta,
 \qquad
 \rho_\eta(x)=\overline\rho_\eta(x)
 \quad\text{for every }x\in\mathcal X.
 \label{eq:compatible_selector_exactness_new}
\end{equation}
\end{proposition}

\begin{proof}
Take
$c\in\mathcal U_{\mathrm{SPO+}}(\hat c;\eta)$ and let
$v_k=x^*(c)$ be the vertex returned by the fixed selector.  Its optimality
gives
$c^\top(v_k-v_j)\leq0$ for every $j$.  The score inequality is equivalent
to
\[
 (c-2\hat c)^\top(v_j-v_k)\leq\eta
 \qquad\text{for every }j,
\]
which is the upper family defining $P_k$.  Hence $c\in P_k$, proving
\eqref{eq:tie_safe_set_containment_new}.  If the optimizer under $c$ is
unique, any $c\in P_k$ must use that same $v_k$ in the score, so the upper
piece inequalities also prove the reverse inclusion.

The selector-specific set contains
$\mathcal U_\eta^{\mathrm{uni}}$ and is contained in
$\overline{\mathcal U}_\eta$.  Taking closures and applying
\eqref{eq:unique_region_closure_new} proves
\eqref{eq:tie_safe_closure_new}.

For any $c\in P_k$, the lower inequalities make $v_k$ optimal even if
other vertices are tied.  Regret is invariant to which optimizer is used,
and therefore
\begin{equation}
 \operatorname{Reg}(x,c)=c^\top(x-v_k)
 \qquad\text{for every }x\in\mathcal X.
 \label{eq:tie_piece_regret_identity_new}
\end{equation}
The regret is continuous in $c$, so
\eqref{eq:tie_safe_closure_new},
\eqref{eq:tie_piece_regret_identity_new}, and
Proposition~\ref{prop:finite_piece_regret} give
\eqref{eq:tie_safe_rho_order_new}.

It remains to prove exactness of the compatible selector.  For a fixed
$c$, all its optimal vertices share the same value
$m_c:=\min_jc^\top v_j$.  Let
\[
 M(c):=\max_j(c-2\hat c)^\top v_j.
\]
If $v$ is any $c$-optimal vertex, the SPO+ score obtained by using $v$ is
\begin{equation}
 M(c)-(c-2\hat c)^\top v
 =M(c)-m_c+2\hat c^\top v.
 \label{eq:tied_optimizer_score_order_new}
\end{equation}
The compatible selector minimizes the final term among all $c$-optimal
vertices and therefore gives the smallest possible tied-optimizer score.
Now take $c\in P_k$.  The vertex $v_k$ is $c$-optimal and its upper piece
inequalities say that its score is at most $\eta$.  By
\eqref{eq:tied_optimizer_score_order_new}, the compatible selector has a
score no larger than this one, so
$c\in\mathcal U_{\mathrm{SPO+}}(\hat c;\eta)$.  This proves the reverse
inclusion and hence the set equality in
\eqref{eq:compatible_selector_exactness_new}.  The equality of robust
values then follows from~\eqref{eq:tie_piece_regret_identity_new}.
\end{proof}

For completeness, Proposition~\ref{prop:tie_safe_extension} also completes the arguments concerning
attainment and probability statements under an arbitrary selector.  By
\eqref{eq:tie_safe_rho_order_new}, the selector-specific $\rho_\eta$ is the
finite maximum of the continuous functions $\phi_k$, and hence is
continuous.  It therefore attains a minimum on compact $\mathcal X$, while
attainment over finite $\mathcal V$ is immediate.
Furthermore,
\[
 \rho_\eta(\hat v)
 =\overline\rho_\eta(\hat v)
 \leq(\eta-2\Gamma)_+.
\]
The nested-domain ordering in Theorem~\ref{thm:margin_refinement} therefore
holds for the selector-specific refinements as well.  On the conformal
event, the realized cost belongs to the selector-specific score set by
definition, so its regret is bounded directly by the corresponding
$\rho_\eta$.  Under any deterministic selector, the piecewise
representations, truncation bounds, and continuous master compute the exact
worst-case value.  A closed-piece maximizer may nevertheless be merely a
limit point of the selector-specific set; the compatible selector
additionally makes the underlying set equality and closed-piece adversarial
attainment exact.

\paragraph{Nominal ties, equality boundaries, and singleton polytopes.}
Continue to use the closed-union notation introduced above.
Suppose now that $\hat v$ is any fixed vertex selected from a non-singleton
optimal face under $\hat c$.  Then $\Delta_k\geq0$ still holds for every
vertex, and all vertices of the nominal optimal face have zero gap.  The
graph of a face of a polytope is connected.  From $\hat v$, first traverse
zero-gap edges inside the nominal face as follows.  For a target vertex
$v$, apply the decreasing-edge argument from the proof of
Proposition~\ref{prop:active_connectivity} until it terminates at some
optimal vertex $w$.  The nominal face contains both $\hat v$ and $w$, and
its one-skeleton gives a zero-gap path from $\hat v$ to $w$.  Concatenating
this path with the reversal of the descent from $v$ to $w$ constructs a
nondecreasing path from the selected $\hat v$ to $v$.  Hence the induced
sublevel graph remains connected without nominal uniqueness.

If the nominal face contains another vertex, the first edge of a path to
that vertex has zero gap.  Consequently,
\[
 \Gamma
 =\min_{v\in\mathcal V\setminus\{\hat v\}}
   \hat c^\top(v-\hat v)
 =\min_{v\in\operatorname{Adj}(\hat v)}
   \hat c^\top(v-\hat v)
 =0.
\]
The margin certificate remains valid but reduces to the unstrengthened
bound $\overline\rho_\eta(\hat v)\leq\eta$.

The inclusive screening boundary is also substantive.  If
$\Delta_k=\eta/2$, the witness $c=0$ keeps $P_k$ nonempty, while
\eqref{eq:screening_necessity_chain_new} forces
$c^\top(\hat v-v_k)=0$ throughout the piece.  Such a piece may be
lower-dimensional.  For example, with
$\mathcal X=[0,1]$, $\hat c=1$, $\hat v=0$, $\eta=2$, and $v_k=1$,
its constraints reduce to $0\leq-c\leq0$, so $P_k=\{0\}$.
Likewise, at $\eta=2\Gamma<+\infty$, closest nonnominal pieces are active
but~\eqref{eq:nonnominal_margin_chain_new} forces their regret at
$\hat v$ to be zero.  This is why the zero-certificate condition is
inclusive.

If $|\mathcal V|=1$, define $\Gamma:=+\infty$ and, for finite $\eta$,
$(\eta-2\Gamma)_+:=0$.  There is no nonnominal or adjacent vertex, so the
local-neighbor formula is not invoked.  Every feasible decision equals
$\hat v$, every regret and refinement value is zero, and all margin
conclusions hold under this convention.  If a pure or continuous
refinement has multiple minimizers in a nonsingleton problem, every
minimizer has the same certificate value; a deterministic measurable tie
rule is needed only to make the selected decision a well-defined random
variable.

\paragraph{Context-dependent feasible sets and nonpositive costs.}
The rank argument remains valid when the feasible polytope varies across
observations, provided the full observations
\[
 (Z_i,A_i,b_i,C_i),\qquad i=1,\ldots,N+1,
\]
are exchangeable, the training data are independent of this joint tuple,
and every realized
$\mathcal X_i=\{x:A_ix=b_i,\ x\geq0\}$ is nonempty and compact.  Assume
also that the same jointly measurable optimization and tie-selection maps
are used to construct every calibration and test score.  Conditional on
training, the resulting scores are then exchangeable.  All geometric proofs
are applied pointwise to each realized polytope.  This formulation includes
instance-specific prices and capacities in the fractional-knapsack
experiments after the usual slack-variable standard-form lifting.

\begin{corollary}[Validity with the nonpositive-cost support]
\label{cor:nonpositive_support_extension}
Assume in addition that $C\in\mathbb R_-^n$ almost surely and replace every
closed piece by $P_k\cap\mathbb R_-^n$.  For finite $\eta$, exact screening
of these supported pieces, finiteness and attainment of piecewise regret, the truncation
bound, and the margin certificate remain valid.  On the SPO+ conformal
event, the realized regret is bounded by the corresponding supported
closed-union certificate with probability at least $1-\alpha$.
\end{corollary}

\begin{proof}
The screening necessity argument is unchanged because a supported piece is
a subset of $P_k$.  For sufficiency, the witness $c=0$ used in
\eqref{eq:screening_zero_witness_new} also belongs to
$\mathbb R_-^n$, so
\[
 P_k\cap\mathbb R_-^n\neq\varnothing
 \quad\Longleftrightarrow\quad
 \Delta_k\leq\eta/2.
\]
For every nonempty intersection of these polyhedra,
\[
 \operatorname{rec}(P_k\cap\mathbb R_-^n)
 =\operatorname{rec}(P_k)\cap\mathbb R_-^n
 \subseteq\operatorname{span}(\mathcal X-\mathcal X)^\perp.
\]
Thus every regret direction $x-v_k$ remains orthogonal to the recession
cone, and the finite-generator proof of finiteness and attainment applies
unchanged.  Intersecting a piece can only decrease its support value, so
all truncation and margin upper bounds remain valid.

On the conformal score event, Proposition~\ref{prop:tie_safe_extension}
places $C$ in a closed piece.  The almost-sure sign condition also gives
$C\in\mathbb R_-^n$, so $C$ belongs to the corresponding supported piece
and its realized regret is bounded by the supported closed-union maximum.
The event has probability at least $1-\alpha$.  In the experiments this
support condition follows from the construction $C=-V$ with $V\geq0$.
In the fixed-piece oracle, the same intersection is implemented simply by
adding $c\leq0$; all other duality constraints are unchanged.
\end{proof}

\section{Experiment Details}
\label{app:experiment_details}

\subsection{Data-Generating Process and Split Protocol}
\label{app:knapsack_data_generation}

All fractional-knapsack experiments use the feasible region
\[
\mathcal X(p,b)=\{x\in[0,1]^n:p^\top x\le b\},
\]
where $x_j$ is the fraction of item $j$ selected, $p_j$ is its price, and $b$ is the budget.
We use a global seed of 42 and deterministic child random streams for the signal matrix and observations.  We report one fixed split rather than averages over multiple top-level seeds.

For context dimension $d_z$, a shared signal matrix $\Theta\in\mathbb R_+^{n\times d_z}$ is generated once for each experimental configuration.  Its final two columns are zero. For the remaining columns, we generate a signal-inclusion indicator $D_{j\ell}$ and a coefficient magnitude $A_{j\ell}$ and set
\[
\begin{aligned}
\Theta_{j\ell}&=D_{j\ell}A_{j\ell},\\
D_{j\ell}&\sim\operatorname{Bernoulli}(0.5),\\
A_{j\ell}&\sim\operatorname{Unif}[0.8,1.2].
\end{aligned}
\]
The same $\Theta$ is used for the training, calibration, and test observations. The context $Z$, conditional mean item value $\mu_j(Z)$, realized value $V_j$, and multiplicative noise $E_j$ satisfy
\[
Z\sim\operatorname{Unif}([0,4]^{d_z}),\qquad
\mu_j(Z)=(Z^\top\Theta_{j\cdot})^2,
\]
\[
V_j=\mu_j(Z)E_j,\qquad
E_j\overset{\mathrm{iid}}{\sim}\operatorname{Unif}[0.8,1.2].
\]
The optimization models use the minimization-cost convention $C=-V$.

Each item price is sampled independently and uniformly from the integers 100 through 1000. Let $S=\sum_jp_j$ be the total price and $M=\max_jp_j$ the largest item price. With independent $U_1,U_2\sim\operatorname{Unif}[0,1]$, define an upper budget limit $\bar b$ and sample $b$ as follows:
\[
\begin{aligned}
\bar b&=\min\{S-10^{-9},\\
&\quad\max\{M+10^{-9},S-U_1M\}\},\\
b&=M+U_2(\bar b-M).
\end{aligned}
\]
A single deterministic observation stream is generated and then partitioned into disjoint training, calibration, and test subsets.  All predictive models are trained on values.  Predictions are transformed back to the original value scale and clipped below at zero before conversion to costs.  No finite lower bound inferred from the synthetic value distribution is supplied to an uncertainty set.

\subsection{Calibration and Evaluation Definitions}
\label{app:calibration_evaluation}

For calibration scores $S_1,\ldots,S_{n_{\mathrm{cal}}}$ and miscoverage level $\alpha$, the split-conformal threshold is the exact order statistic
\[
k_\alpha=\left\lceil(n_{\mathrm{cal}}+1)(1-\alpha)\right\rceil,
\qquad
\eta_\alpha=S_{(k_\alpha)}.
\]
Thus, we do not substitute an interpolated empirical quantile for the conformal order statistic.

Let $v(C)\in\arg\min_{u\in\mathcal X(p,b)}C^\top u$.  The SPO$+$ calibration score is
\[
s_+(C,\widehat C)
=
\max_{u\in\mathcal X(p,b)}
(C-2\widehat C)^\top\{u-v(C)\}.
\]
Nominal, hindsight, and SPO$+$ score problems are solved as linear programs.  Optimal-solution ties use a deterministic secondary objective so that the returned solution is a reproducible polytope vertex.

Reported regret uses the equivalent value convention
\[
R(x;V,p,b)
=
\max_{u\in\mathcal X(p,b)}V^\top u-V^\top x.
\]
For test regrets $r_1,\ldots,r_{n_{\mathrm{test}}}$, the performance
summaries are the arithmetic mean, the empirical $0.95$ quantile, and
\[
\operatorname{CVaR}_{0.95}
=
\min_{t\in\mathbb R}
\left\{
t+\frac{1}{0.05n_{\mathrm{test}}}
\sum_{i=1}^{n_{\mathrm{test}}}(r_i-t)_+
\right\},
\]
which assigns exactly $5\%$ empirical probability mass to the upper tail,
using fractional weight at a tied boundary when necessary.  A decision is
recorded as changed when its $\ell_\infty$ distance from the nominal PTO
decision exceeds $10^{-6}$.

\subsection{Illustrative Objective-Space Geometry}

The geometry figure is illustrative and is not constructed from a calibration sample.  Its action set contains the eight unit-circle points
\[
v_k=\left(\cos(k\pi/4),\sin(k\pi/4)\right),
\qquad k=0,\ldots,7.
\]
The predicted cost is $\widehat C=(-1,-0.35)$.  Scores are evaluated on a $241\times241$ grid over $[-2.75,2.75]^2$, using the illustrative values $\eta=1.10$ and residual radius $0.70$.  The four panels show the closed normal-fan regions, the residual set $\{C:\|C-\widehat C\|_2\le0.70\}$, the exact-regret sublevel set, and the SPO$+$ sublevel set.  In the last panel, a normal-fan piece is active precisely when
\[
\widehat C^\top(v_k-\widehat v)\le\eta/2.
\]
These numerical values serve only to visualize the different objective-space geometries.

\subsection{Active Pieces and Margin Early Exit}

The exact-validation configuration is
\[
\begin{aligned}
n&=8,\qquad d_z=8,\\
n_{\mathrm{train}}&=2000,\quad n_{\mathrm{cal}}=1000,\\
n_{\mathrm{test}}&=500,
\end{aligned}
\]
with $\alpha\in\{0.05,0.10,0.20,0.30,0.40\}$.  The predictor is kernel ridge regression with an RBF kernel, regularization $0.1$, and kernel parameter $0.03$.  Both features and multi-output targets are standardized using training data only.

\subsubsection{Exact Enumeration Audit}

For every test context, all vertices of the fractional-knapsack polytope are explicitly enumerated.  The complete active set obtained by enumeration is compared with the adjacency-based search.  The local margin
\[
\Gamma=\min_{v\in\operatorname{Adj}(\widehat v)}\widehat C^\top(v-\widehat v)
\]
is also compared with complete enumeration.  No active-piece cap is imposed in this experiment.

The histogram uses complete active-piece counts.  The margin diagnostic uses $\alpha=0.10$, places $(\eta-2\Gamma)_+$ on the horizontal axis and the nominal set-relative certificate on the vertical axis, and records the frequency with which the margin rule accepts the nominal decision without solving a robust master.

\subsection{Certified Anytime Active-Piece Generation}
\label{app:anytime_piece_generation}

We empirically examine the certified early-stopping rule in Corollary~\ref{cor:gap-ordered-truncation} under the main $n=20$, $\alpha=0.10$ setting.  The predictor, calibration threshold, and $500$ test contexts are held fixed.  The feasible polytopes contain $1.159\times 10^6$ vertices on average, while the complete active sets contain only $44.956$; the full local traversal evaluates $380.472$ distinct vertices on average.  For each context, we enumerate active vertices in nondecreasing predicted gap and consider requested prefix budgets
\[
M\in\{1,5,10,25,50,100\}.
\]
If the $M$th vertex lies on a predicted-gap plateau, the prefix is expanded through the entire plateau. Thus the retained set is exactly
$\mathcal A_\tau=\{k:\Delta_k\leq\tau\}$ for the recorded boundary $\tau$; we denote its size by $M_{\mathrm{eff}}$.  This tie-safe convention altered only two of the $500$ contexts at one grid budget each, changing $M_{\mathrm{eff}}$ from $10$ to $11$ and from $25$ to $27$.

For each tie-safe prefix and each decision domain, we solve the restricted refinement and record
\[
\begin{aligned}
R_\tau&=\rho_{\eta,\tau}(x_\tau),&
B_\tau&=\bigl(\eta+2\widehat C^\top(x_\tau-\widehat v)-2\tau\bigr)_+,\\
U_\tau&=\max\{R_\tau,B_\tau\}.&&
\end{aligned}
\]
We declare exact certification only when $B_\tau\leq R_\tau$ up to the
prespecified scale-aware tolerance, or when the complete active set has been exhausted.  The reported optimality gap is zero in either certified case and $U_\tau-R_\tau$ otherwise, normalized by $\eta$.  As an independent check, every row satisfies $R_\tau\leq R_{\mathcal D}^\star\leq U_\tau$.  The reference $R_{\mathcal D}^\star$ is the full-set certificate established in the main experiment: $472$ contexts are exhausted within $100$ pieces, and the separate full-set audit verifies the remaining $28$ contexts.

Figure~\ref{fig:anytime_certification}(a) shows that exact certification arrives substantially before exhaustive enumeration.  With $M=25$, the certification rates are $99.6\%$ for Pure RR and $91.4\%$ for Continuous RR, whereas only $56.2\%$ of the active sets have been fully enumerated.  At $M=50$, these rates rise to $99.8\%$ and $98.4\%$, respectively; at $M=100$, they are $100\%$ and $99.8\%$.  Equivalently, the first requested grid budget certifying exactness is at most $25$ for $498$ Pure RR instances and $457$ Continuous RR instances.  These are first certifying points on the chosen grid, not theoretical minimum prefix sizes.

As shown in Figure~\ref{fig:anytime_certification}(b), the certified gap contracts rapidly even before exact termination.  For Pure RR, its median is already zero at $M=5$, while its $0.90$ quantile falls from $0.700$ at $M=5$ to $0.128$ at $M=10$ and zero at $M=25$.  For Continuous RR, the corresponding median/$0.90$-quantile values are $0.314/0.739$, $0.030/0.395$, and $0/0$.  Hence the gap bound remains informative in the instances that are not yet exactly certified.

Table~\ref{tab:anytime_summary} separates the cost of prefix generation from the robust master.  The best-first search takes a median of $15.1$ ms and a $0.90$ quantile of $51.8$ ms by $M=100$, whereas the paired master solve has corresponding times of $1.045$ and $2.959$ seconds.  Moreover, the median $M_{\mathrm{eff}}$ saturates at $22$, because more than half of the active sets are already complete.  These results support the output-sensitive interpretation of the method on this benchmark: a small gap-ordered prefix usually suffices for a rigorous full-problem certificate, and the search itself is not the computational bottleneck.

The single Continuous RR instance not certified by $M=100$ is a $4{,}770$-piece extreme case; its normalized certified gap is $0.111$. Thus, rather than recommending $M=100$ as a general rule, the experiment further supports our conclusion that the number of active pieces required in practice is typically small relative to the total number of vertices.

\subsection{Coverage of Calibrated Sets}

\label{app:set_coverage_comparison}

We compare what the calibrated sets cover in parameter space with what their induced certificates cover at the decision level.  This experiment uses the same $n=8$ split, RBF kernel-ridge predictor, and five calibration levels as the exact-validation experiment.  For method $m$, let $\mathcal U_{m,\alpha}(Z_i)$ denote its calibrated uncertainty set and let $x_{m,\alpha,i}$ be the decision returned from that set.  We record
\[
I^{\mathrm{par}}_{m,\alpha,i}
=\mathbf 1\{C_i\in\mathcal U_{m,\alpha}(Z_i)\}
\]
and
\[
\begin{aligned}
I^{\mathrm{reg}}_{m,\alpha,i}
&=\mathbf 1\!\left\{
\operatorname{Reg}(x_{m,\alpha,i},C_i)
\le \rho_{m,\alpha,i}\right\},\\
\rho_{m,\alpha,i}
&=\sup_{C'\in\mathcal U_{m,\alpha}(Z_i)}
\operatorname{Reg}(x_{m,\alpha,i},C').
\end{aligned}
\]
The SPO$+$ row uses the continuous regret-refinement decision.  The other three rows use their method-specific Cost-RO decisions, after which $\rho_{m,\alpha,i}$ is evaluated over the complete corresponding uncertainty set. Thus, the experiment compares the certificates naturally attached to the decisions each method actually prescribes; it does not replace the Cost-RO objectives of the conventional baselines by regret minimization.

\begin{table*}[ht]
\centering
\small
\setlength{\tabcolsep}{4pt}
\caption{Certified anytime behavior by requested prefix budget.  Exact, full-set, and tie-expansion columns are percentages; effective prefix sizes and runtimes are reported as median / $0.90$ quantile. Search time is in milliseconds and paired-solve time is in seconds.}
\label{tab:anytime_summary}
\begin{tabular}{@{}lrrrrrrr@{}}
\toprule
Domain & $M$ & Exact & Full set & Tie exp. &
$M_{\mathrm{eff}}$ & Search (ms) & Solve (s) \\
\midrule
Pure RR & 1   & 0.8   & 0.8  & 0.0 & 1.0/1.0   & 0.769/1.231   & 0.049/0.108 \\
Pure RR & 5   & 63.2  & 5.8  & 0.0 & 5.0/5.0   & 3.418/4.577   & 0.312/0.606 \\
Pure RR & 10  & 86.6  & 21.6 & 0.2 & 10.0/10.0 & 6.668/8.487   & 0.483/1.045 \\
Pure RR & 25  & 99.6  & 56.2 & 0.2 & 22.0/25.0 & 14.931/18.760 & 0.833/1.557 \\
Pure RR & 50  & 99.8  & 83.2 & 0.0 & 22.0/50.0 & 15.055/33.920 & 1.045/2.107 \\
Pure RR & 100 & 100.0 & 94.4 & 0.0 & 22.0/72.1 & 15.055/51.759 & 1.045/2.959 \\
\midrule
Continuous RR & 1   & 0.8  & 0.8  & 0.0 & 1.0/1.0   & 0.769/1.231   & 0.049/0.108 \\
Continuous RR & 5   & 17.8 & 5.8  & 0.0 & 5.0/5.0   & 3.418/4.577   & 0.312/0.606 \\
Continuous RR & 10  & 46.4 & 21.6 & 0.2 & 10.0/10.0 & 6.668/8.487   & 0.483/1.045 \\
Continuous RR & 25  & 91.4 & 56.2 & 0.2 & 22.0/25.0 & 14.931/18.760 & 0.833/1.557 \\
Continuous RR & 50  & 98.4 & 83.2 & 0.0 & 22.0/50.0 & 15.055/33.920 & 1.045/2.107 \\
Continuous RR & 100 & 99.8 & 94.4 & 0.0 & 22.0/72.1 & 15.055/51.759 & 1.045/2.959 \\
\bottomrule
\end{tabular}
\end{table*}

\begin{table*}[!t]
\centering
\small
\setlength{\tabcolsep}{5pt}
\caption{Marginal parameter and decision-regret coverage on the fixed
$n=8$ test split.  Each entry is parameter coverage / decision-regret
coverage; the nominal target is $1-\alpha$.}
\label{tab:set_coverage_comparison}
\begin{tabular}{@{}lccccc@{}}
\toprule
Method & $\alpha=.05$ & $\alpha=.10$ & $\alpha=.20$ & $\alpha=.30$ & $\alpha=.40$ \\
\midrule
SPO$+$ regret set (Continuous RR)
& 0.964/0.974 & 0.912/0.926 & 0.826/0.862 & 0.706/0.802 & 0.656/0.816 \\
Residual-$\ell_2$ Cost-RO
& 0.938/0.998 & 0.892/0.998 & 0.804/0.998 & 0.696/0.996 & 0.572/0.992 \\
Contextual Box Cost-RO
& 0.942/1.000 & 0.904/1.000 & 0.796/1.000 & 0.714/1.000 & 0.608/0.998 \\
KNN--Ellipsoid Cost-RO
& 0.936/1.000 & 0.868/1.000 & 0.778/1.000 & 0.678/1.000 & 0.570/1.000 \\
\bottomrule
\end{tabular}
\end{table*}

As shown in Table~\ref{tab:set_coverage_comparison}, parameter coverage generally follows the nominal sequence, with the modest departures expected from one fixed calibration--test split.  The largest Bernoulli standard error for $500$ test observations is $0.022$, so small differences from $1-\alpha$ should not be interpreted as evidence for or against the marginal guarantee.  More importantly, parameter inclusion deterministically implies decision-regret coverage for every method, but the converse need not hold.
The conventional sets consequently attain decision-regret coverage between $0.992$ and $1.000$, even when their parameter coverage is close to the nominal target.  The corresponding gap is smaller for the SPO$+$ set, especially at the lower miscoverage levels.  High decision-regret coverage alone, however, may result from a loose certificate rather than a more informative characterization of decision risk.

To distinguish these explanations, Table~\ref{tab:set_certificate_scale} reports the scale of the same full-set certificates.  The SPO$+$ certificate has a median of $4.990$ at $\alpha=.05$ and $1.407$ at $\alpha=.10$, and its median is exactly zero from $\alpha=.20$ onward.  Its $0.95$ quantile falls from $7.697$ to zero across the calibration levels.  By contrast, the Residual-$\ell_2$ and KNN--Ellipsoid certificates remain much larger throughout.  The contextual box has a slightly smaller median at $\alpha=.05$, but its $0.95$ quantile is $28.944$, compared with $7.697$ for SPO$+$.  Hence, the near-universal decision-regret coverage of the conventional sets is accompanied by substantially wider regret bounds.  This comparison supports the distinction motivating our construction: comparable marginal coverage in parameter space does not by itself imply a decision-informative certificate, whereas calibration in the SPO$+$ geometry more closely couples coverage to the downstream regret scale. These comparisons concern marginal coverage; they do not assert conditional coverage or uniform dominance in realized decision performance.

\begin{table*}[!t]
\centering
\small
\setlength{\tabcolsep}{5pt}
\caption{Scale of the method-specific full-set regret certificates in
Table~\ref{tab:set_coverage_comparison}.  Each entry is the median /
$0.95$ quantile over the $500$ test observations.}
\label{tab:set_certificate_scale}
\begin{tabular}{@{}lccccc@{}}
\toprule
Method & $\alpha=.05$ & $\alpha=.10$ & $\alpha=.20$ & $\alpha=.30$ & $\alpha=.40$ \\
\midrule
SPO$+$ regret set (Continuous RR)
& 4.990/7.697 & 1.407/3.031 & 0.000/0.751 & 0.000/0.073 & 0.000/0.000 \\
Residual-$\ell_2$ Cost-RO
& 35.783/55.768 & 30.142/47.248 & 23.427/37.042 & 18.935/30.515 & 14.934/25.445 \\
Contextual Box Cost-RO
& 4.033/28.944 & 3.531/26.607 & 2.792/21.779 & 2.488/20.421 & 2.132/18.512 \\
KNN--Ellipsoid Cost-RO
& 88.626/244.809 & 50.288/165.904 & 28.770/90.569 & 20.315/60.454 & 15.750/46.285 \\
\bottomrule
\end{tabular}
\end{table*}

\subsection{Main Decision-Performance Comparison}
\label{app:main_performance_setup}

\subsubsection{Experimental Setup and Refinement Implementation}

The main comparison uses
\[
\begin{aligned}
n&=20,\qquad d_z=15,\qquad\alpha=0.10,\\
n_{\mathrm{train}}&=5000,\quad n_{\mathrm{cal}}=2000,\\
n_{\mathrm{test}}&=500.
\end{aligned}
\]
All methods use the same observations and test contexts.  The principal predictor is polynomial kernel ridge regression, selected by training-only five-fold cross-validation. The tuning grid contains ten logarithmically spaced regularization strengths $\lambda$ from $10^{-3}$ to $10^3$ and polynomial degrees $d\in\{2,3,4\}$. The kernel scale is $\gamma=1/d_z$, and the offset is $\mathrm{coef0}=1$.  The selected configuration is degree $2$ with $\lambda=2.154$; its mean cross-validation MSE is $999.114$.  Feature and multi-output target standardization are fitted inside each training fold, after which the selected model is refitted on all $5000$ training observations.

The compared methods are Nominal PTO, SPO$+$ Pure RR, SPO$+$ Continuous RR, Conformal Residual-$\ell_2$ Cost-RO, Conformal Box Cost-RO, and KNN--Ellipsoid Cost-RO.  The two SPO$+$ methods minimize worst-case regret over the calibrated SPO$+$ set.  Pure RR restricts the returned decision to a vertex of the fractional-knapsack polytope, whereas Continuous RR optimizes over the full fractional polytope.

The active-piece search starts at the nominal vertex and uses a priority queue ordered by the predicted objective gap
\[
\delta(v)=\widehat C^\top(v-\widehat v).
\]
Canonical vertex keys, rounded to ten decimal places, break equal-gap ties.  Only vertices satisfying
\[
\delta(v)\le\eta/2+10^{-9}\max\{1,|\eta|\}
\]
are expanded.  For computational efficiency, the pure and continuous masters are first solved using the same best-first prefix of at most $100$ active pieces.  Every returned decision is then reevaluated against the complete active set, so the reported certificates use the full set rather than the retained prefix.  The complete set contains at most $100$ pieces in $472$ of the $500$ test contexts.  In each of the remaining $28$ contexts, the full-set certificate is exactly equal to its prefix certificate for both refinements; hence the prefix solution is also optimal for the full active set.  The worst-case pieces have maximum prefix ranks $28$ and $27$ for Pure and Continuous RR, respectively.  Thus, the $100$-piece limit is only a computational device and does not alter any reported decision or certificate.  When $\eta\le2\Gamma$, up to the scale-aware numerical tolerance, the nominal decision is returned directly.

\subsubsection{Cost-RO Baselines}

Unlike the proposed refinements, the three conventional baselines solve
\[
\min_{x\in\mathcal X(p,b)}
\max_{C\in\mathcal U(Z)}C^\top x
\]
and do not optimize worst-case regret. Each baseline has its own calibration score and conformal threshold. Since item values are nonnegative, the minimization costs $C=-V$ satisfy $C_j\le0$. For solver implementation, strict negativity is represented numerically by
\[
C_0^\epsilon=\{C:C_j\le-\epsilon_s,\ j=1,\ldots,n\},
\qquad
\epsilon_s=10^{-6}.
\]
Replacing the boundary value $0$ with $-\epsilon_s$ is primarily intended to improve numerical stability while providing a solver-compatible implementation of $C_j<0$. The SPO$+$ fixed-piece models use the closed limiting support $C_j\le0$.

For Residual-$\ell_2$ Cost-RO, the score and calibrated set are
\[
\begin{aligned}
S_{\ell_2}(C,\widehat C)
&=\|C-\widehat C\|_2,\\
\mathcal U_{\ell_2}
&=\{C:\|C-\widehat C\|_2\le\eta_{\ell_2}\}\\
&\quad\cap C_0^\epsilon.
\end{aligned}
\]
The direct SOCP reformulation uses dual variables $\lambda\in\mathbb R_+^n$ for the cost-support constraints and solves
\[
\min_{x,\lambda}
\widehat C^\top(x-\lambda)
+\eta_{\ell_2}\|x-\lambda\|_2
-\epsilon_s\mathbf 1^\top\lambda
\quad\text{s.t. }x\in\mathcal X(p,b).
\]

For the contextual Box baseline, a separate gradient-boosting quantile regressor is fitted to the absolute training residuals of each item.  Its settings are \texttt{loss=quantile}, \texttt{alpha=0.8}, \texttt{n\_estimators=40}, \texttt{max\_depth=2}, and \texttt{random\_state=42}.
Its predicted scale is truncated below at $10^{-6}$.  With scale vector $H(Z)$, the calibration score is
\[
S_{\mathrm{Box}}(C,\widehat C,Z)
=
\max_j\frac{|C_j-\widehat C_j|}{H_j(Z)}.
\]
The calibrated coordinate bounds are
\[
\widehat C_j-\eta_{\mathrm{Box}} H_j(Z)
\le C_j\le
\widehat C_j+\eta_{\mathrm{Box}} H_j(Z),
\]
intersected with $C_0^\epsilon$.  Since $x\ge0$, the robust problem is the LP
\[
\min_{x\in\mathcal X(p,b)}
\sum_j
\min\{\widehat C_j+\eta_{\mathrm{Box}} H_j(Z),-\epsilon_s\}x_j.
\]

For KNN--Ellipsoid Cost-RO, features are standardized using the training split.  Each calibration or test context uses its $k=100$ nearest training observations.  Their observed costs define a local mean $\mu_K(Z)$; the local covariance $\Sigma_K(Z)$ is estimated using Ledoit--Wolf shrinkage.  If its smallest eigenvalue is below $10^{-10}$, the implementation adds the following multiple of the $n\times n$ identity matrix $I$:
\[
10^{-8}\frac{\operatorname{tr}(\Sigma_K)}{n}I.
\]
The calibration score and uncertainty set are
\[
S_K(Z,C)
=
\sqrt{(C-\mu_K)^\top\Sigma_K^{-1}(C-\mu_K)},
\]
\[
\mathcal U_K(Z)
=
\{C:(C-\mu_K)^\top\Sigma_K^{-1}(C-\mu_K)\le\eta_K^2\}
\cap C_0^\epsilon.
\]
With covariance factor $L$ satisfying $\Sigma_K=LL^\top$, the direct SOCP reformulation is
\[
\begin{aligned}
\min_{x,\lambda\ge0}\quad&
\mu_K^\top(x-\lambda)
+\eta_K\|L^\top(x-\lambda)\|_2\\
&-\epsilon_s\mathbf 1^\top\lambda\\
\text{s.t.}\quad&x\in\mathcal X(p,b).
\end{aligned}
\]

\subsubsection{Regret-Distribution Profiles}
\label{app:regret_distribution_profiles}

\begin{figure}[htb]
    \centering
    \includegraphics[width=0.8\linewidth]{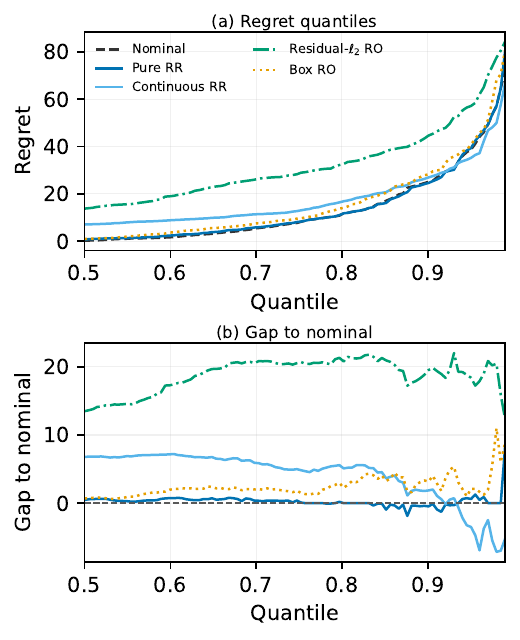}
    \caption{Realized-regret distribution at $\alpha=0.10$.  (a) Empirical regret quantiles.  (b) Quantile-wise gap to Nominal PTO, where negative values are better.  KNN--Ellipsoid is omitted for scale and retained in Table~\ref{tab:performance_main}.}
    \label{fig:performance_quantiles}
\end{figure}

The summary statistics in Table~\ref{tab:performance_main} do not identify where the tail benefit of refinement begins. We therefore compare the empirical regret distributions at $\alpha=0.10$ over $100$ evenly spaced quantile levels from $0.50$ through $0.99$.  Figure~\ref{fig:performance_quantiles}(a) displays the quantile curves, while panel (b) subtracts the corresponding Nominal PTO quantile to make the location and direction of each difference explicit.

Continuous RR performs worse over the center of the distribution but improves upon Nominal PTO from approximately the $0.94$ quantile onward, thereby reducing severe regret.  Pure RR closely follows the nominal curve and intervenes only in the subset of contexts identified as vulnerable by the calibrated regret model.  Hence, continuous refinement emphasizes tail protection, while pure refinement adds robustness with limited disruption to nominal performance.

\subsection{Objective-Matched Regret Comparison}
\label{app:objective_matched}

We complement the main comparison with minimax-regret variants of the three residual-based RO baselines. The experiment uses the same $n=20$ fractional-knapsack instances, predictor, calibration sample, and $500$ test observations at $\alpha=0.10$ as Table~\ref{tab:performance_main}. For each calibrated set $\mathcal U_m(Z)$, where $m$ denotes Residual-$\ell_2$, Box, or KNN--Ellipsoid, we solve
\begin{equation}
\min_{x\in\mathcal X(p,b)}
\max_{c\in\mathcal U_m(Z)}
\left\{c^\top x-\min_{y\in\mathcal X(p,b)}c^\top y\right\}.
\label{eq:objective_matched_rr}
\end{equation}
Each baseline retains its original calibrated set and support constraints, changing only the downstream objective. The decision domain is the full fractional polytope, as in Continuous RR. All three baseline variants attain a relative optimality gap of at most $10^{-6}$ on all $500$ test instances.

\begin{table}[htb]
\centering
\footnotesize
\setlength{\tabcolsep}{2.5pt}
\caption{Objective-matched performance at $n=20$ and $\alpha=0.10$ on $500$ test observations. Change is the decision-change rate; Time is mean recorded time per instance (one solver thread for PTO, SPO+, and Box; up to twenty for Residual-$\ell_2$ and KNN).}
\label{tab:objective_matched}
\begin{tabular}{@{}lrrrrr@{}}
\toprule
Method & Mean & Q95 & CVaR95 & Change & Time (s) \\
\midrule
Nominal PTO & \textbf{7.605} & 38.664 & 60.557 & 0.000 & 0.00124 \\
Pure RR & 7.763 & 38.956 & 61.814 & 0.100 & 1.41800 \\
Continuous RR & 11.117 & \textbf{35.137} & \textbf{52.618} & 0.990 & 1.88969 \\
Box RR & 17.890 & 57.051 & 73.152 & 0.986 & 0.10363 \\
Residual-$\ell_2$ RR & 38.348 & 64.456 & 78.628 & 1.000 & 13.15628 \\
KNN RR & 52.924 & 119.828 & 146.999 & 1.000 & 14.37106 \\
\bottomrule
\end{tabular}
\end{table}

Table~\ref{tab:objective_matched} shows that both SPO+ refinements achieve lower mean and tail regret than all three objective-matched baselines. Switching KNN from worst-case cost to minimax regret improves its mean regret from $624.111$ to $52.924$, while the advantage of SPO+ remains. These results support the contribution of the SPO+-calibrated set beyond the choice of downstream objective.

Pure and Continuous RR are run consecutively with shared active-piece search and model construction; their reported times include this common work and their respective optimization and certification stages.

\subsection{Additional Validation of Marginal Guarantees}

As a supplementary check using the same $n=8$ exact-validation split, we evaluate the prespecified marginal events for the proposed method across all five calibration levels.

For each $\alpha$, we evaluate the six empirical events
\[
s_+(C,\widehat C)\le\eta_\alpha,
\qquad
\operatorname{Reg}(\widehat v,C)\le\eta_\alpha,
\]
\[
\begin{aligned}
\operatorname{Reg}(x^{\mathrm{pure}},C)
&\le\rho_{\eta_\alpha}(x^{\mathrm{pure}}),\\
\operatorname{Reg}(x^{\mathrm{cont}},C)
&\le\rho_{\eta_\alpha}(x^{\mathrm{cont}}).
\end{aligned}
\]
\[
\begin{aligned}
\operatorname{Reg}(x^{\mathrm{pure}},C)
&\le(\eta_\alpha-2\Gamma)_+,\\
\operatorname{Reg}(x^{\mathrm{cont}},C)
&\le(\eta_\alpha-2\Gamma)_+.
\end{aligned}
\]
An event $a\le b$ is accepted numerically when
\[
a\le b+10^{-6}\max\{1,|b|\}.
\]
The validation table reports the empirical frequencies with target $1-\alpha$.

Table~\ref{tab:validation_coverage} reports the six prespecified marginal events.  SPO$+$ score inclusion frequencies are $0.964$, $0.912$, $0.826$, $0.706$, and $0.656$, compared with targets $0.95$, $0.90$, $0.80$, $0.70$, and $0.60$, respectively.  The nominal-regret, set-relative, and margin-certificate events are likewise at or above their corresponding targets. These fixed-seed empirical frequencies are consistent with the marginal guarantees up to ordinary finite-sample variation.
\begin{table*}[t]
\centering
\small
\setlength{\tabcolsep}{5pt}
\caption{Empirical probabilities for the six validation events; nominal target $1-\alpha$. These marginal frequencies summarize one fixed-seed experiment and do not claim conditional coverage.}
\label{tab:validation_coverage}
\begin{tabular}{lccccc}
\toprule
Event & $\alpha=0.05$ & $\alpha=0.10$ & $\alpha=0.20$ & $\alpha=0.30$ & $\alpha=0.40$ \\
\midrule
SPO$+$ score inclusion & 0.964 & 0.912 & 0.826 & 0.706 & 0.656 \\
Nominal regret & 0.994 & 0.978 & 0.928 & 0.850 & 0.816 \\
Pure RR set-relative certificate & 0.978 & 0.944 & 0.882 & 0.818 & 0.816 \\
Continuous RR set-relative certificate & 0.974 & 0.926 & 0.862 & 0.802 & 0.816 \\
Pure RR margin certificate & 0.980 & 0.946 & 0.888 & 0.820 & 0.816 \\
Continuous RR margin certificate & 0.982 & 0.942 & 0.882 & 0.806 & 0.816 \\
\bottomrule
\end{tabular}
\end{table*}

\subsection{Additional Predictor-Quality Comparison}

\subsubsection{Predictors and Model Selection}

This experiment uses the same $n=20$ data, fixed split, test contexts, seed, and $\alpha=0.10$ as the main comparison.  It includes only Nominal PTO, SPO$+$ Pure RR, and SPO$+$ Continuous RR.  Each predictor has its own SPO$+$ calibration scores and conformal threshold; a threshold calibrated for one predictor is never reused for another.

All new predictors are selected by deterministic five-fold cross-validation on the training split.  The folds are shuffled using seed 42, and selection minimizes mean multi-output validation MSE.  Grid search uses four parallel workers, with estimator-level parallelism restricted to one worker.  The selected model is refitted on the complete training split.  Polynomial KRR reuses the model, predictions, and calibration evidence from the main comparison.

\paragraph{Ridge.}
The regularization grid contains nine logarithmically spaced values of $\lambda$ from $10^{-4}$ to $10^4$.  Feature and target standardization are fitted within each fold.  The selected value is $\lambda=1$, with mean cross-validation MSE $1151.114$.

\paragraph{Polynomial KRR.}
The kernel and tuning grid follow Appendix~\ref{app:main_performance_setup}. The reused configuration is $d=2$ and $\lambda=2.154$, with mean cross-validation MSE $999.114$.

\paragraph{RBF KRR.}
The grid is
\[
\begin{aligned}
\lambda&\in\{10^{-3},10^{-2},10^{-1},1,10\},\\
\gamma&\in d_z^{-1}\{0.25,0.5,1,2\}.
\end{aligned}
\]
Features and targets are standardized within each fold.  The selected configuration is $\lambda=1$ and $\gamma=1/60=0.01667$, with mean cross-validation MSE $1033.416$.

\paragraph{MLP.}
The candidate hidden-layer configurations are $(64)$ and $(128,64)$, with weight-decay values $\{10^{-5},10^{-4},10^{-3}\}$ and initial learning rates $\{3\times10^{-4},10^{-3}\}$.  All candidates use ReLU activations, Adam, at most $500$ epochs, an early-stopping validation fraction of $0.1$, a $30$-epoch no-improvement rule, and seed 42.  Features and targets are standardized within each fold.  The selected configuration has one 64-unit hidden layer, weight decay $10^{-3}$, and initial learning rate $10^{-3}$; its mean cross-validation MSE is $1032.420$.

\paragraph{Random Forest.}
The grid is
\[
\begin{aligned}
\texttt{n\_estimators}&\in\{200,400\},\\
\texttt{min\_samples\_leaf}&\in\{1,2,5\},\\
\texttt{max\_features}&\in\{0.5,1.0\}.
\end{aligned}
\]
The model uses raw features and raw multi-output values, seed 42, and one estimator worker.  The selected configuration has 400 trees, $\texttt{min\_samples\_leaf}=1$, and $\texttt{max\_features}=1.0$, with mean cross-validation MSE $3106.874$.

The independently calibrated SPO$+$ thresholds, in the row order of Table~\ref{tab:prediction_quality}, are $110.280$, $88.482$, $90.922$, $91.778$, and $202.496$.  Test RMSE is computed jointly over all test observations and items.  For each predictor, the reported mean active-piece count comes from a complete, uncapped adjacency search.  The Pure and Continuous RR masters subsequently use at most the first 150 active pieces in best-first order. The raw output stores the complete count, used count, and truncation indicator separately.  The margin-skip rate is the fraction of test contexts satisfying $\eta\le2\Gamma$.  No Cost-RO baseline is rerun in this experiment.

\subsubsection{Exploratory Results}
Table~\ref{tab:prediction_quality} summarizes prediction accuracy, mean decision regret, and complete active-piece counts for the five predictors.  The refinement masters use at most the first 150 pieces in best-first order, so this computational cap does not truncate the reported geometric statistic.
\begin{table*}[t]
\centering
\small
\setlength{\tabcolsep}{5pt}
\caption{Prediction quality, decision regret, and refinement geometry. Active counts use the complete uncapped search; Lower values are better except for the skip rate.}
\label{tab:prediction_quality}
\begin{tabular}{@{}lrrrrrr@{}}
\toprule
Predictor & RMSE & Nominal & Pure RR & Continuous RR & Active & Skip (\%) \\
\midrule
Ridge & 33.87 & 10.08 & 9.97 & 14.60 & 370.63 & 0.20 \\
Poly. KRR & 31.55 & 7.60 & 7.76 & 11.12 & 44.96 & 0.80 \\
RBF KRR & 32.15 & 7.56 & 7.70 & 11.35 & 45.47 & 0.80 \\
MLP & 31.96 & 7.75 & 7.84 & 11.74 & 45.66 & 0.60 \\
Random forest & 53.94 & 20.17 & 20.88 & 29.50 & 181.42 & 0.20 \\
\bottomrule
\end{tabular}
\end{table*}

Prediction accuracy and downstream performance are related but not interchangeable.  Polynomial KRR has the lowest test RMSE ($31.55$), whereas RBF KRR has the lowest Nominal and Pure-RR mean regrets ($7.56$ and $7.70$); polynomial KRR instead has the lowest Continuous-RR mean regret ($11.12$).  Pure RR remains close to Nominal PTO for every predictor and slightly improves Ridge from $10.08$ to $9.97$, but is slightly worse for the other four.  Continuous RR has higher mean regret than Nominal PTO throughout, consistent with the mean--tail trade-off observed in the main comparison rather than a uniform improvement in average performance.  Random forest is weakest in both RMSE and regret, although the remaining prediction and decision rankings are not identical.

Geometric complexity is even less closely related to prediction quality.  Polynomial KRR, RBF KRR, and the MLP have similar complete mean active-piece counts ($44.96$, $45.47$, and $45.66$), whereas Random Forest has $181.42$ and Ridge has the largest mean, $370.63$, despite substantially outperforming Random Forest in prediction and decision quality.  Margin certificates trigger on only $0.2\%$--$0.8\%$ of contexts.  The 150-piece refinement cap binds in 232 of the 2,500 predictor--context pairs, including 151 Random-Forest pairs; the reported active counts nevertheless use the complete search.  These fixed-split comparisons are descriptive: the predictor jointly changes its errors, calibrated score distribution, and local decision geometry, so RMSE alone does not determine regret or search burden.

\subsection{Optimization and Numerical Settings}
\label{app:optimization_settings}

The main knapsack experiments use
\[
\begin{aligned}
\texttt{OutputFlag}&=0,\qquad \texttt{Threads}=1,\\
\texttt{Seed}&=42.
\end{aligned}
\]
\[
\begin{aligned}
\texttt{MIPGap}&=10^{-6},\\
\texttt{FeasibilityTol}&=10^{-7},\\
\texttt{OptimalityTol}&=10^{-7}.
\end{aligned}
\]
Fractional-knapsack LPs use deterministic dual simplex and a secondary tie-breaking optimization that preserves the primary objective.  The pure vertex master additionally uses $\texttt{IntFeasTol}=10^{-9}$.  The auxiliary mixed-integer regret oracle used to evaluate the Cost-RO certificates in Table~\ref{tab:set_certificate_scale} uses $\texttt{FeasibilityTol}=10^{-9}$ and $\texttt{IntFeasTol}=10^{-7}$.  The tighter primal tolerance prevents a feasible-vertex decomposition error from being amplified by the cost coefficients.

The SPO$+$ masters use constraint generation with at most 200 iterations and add all currently violated cuts at each iteration.  The final certificate is independently recomputed over all retained fixed-piece oracles.  A solve is accepted only if Gurobi returns \texttt{OPTIMAL}, any applicable relative gap is at most $10^{-6}$, all certificates are finite, and the normalized physical feasibility residual is at most $10^{-6}$.  The continuous, pure, and nominal certificates must also satisfy their expected hierarchy up to a scale-aware tolerance of $10^{-6}$.  Failures are treated as errors; no failed instance is silently replaced by its nominal decision or removed from the test sample.  The Cost-RO decisions themselves use one LP or SOCP and do not optimize through the bilinear adversarial-regret oracle; that oracle is invoked only afterward to evaluate the supplementary full-set certificates.

\subsection{Computing Environment}

All experiments were run on a PC with an Intel Core Ultra 9
275HX processor and 32 GB of RAM, using 64-bit Microsoft Windows 11.
The implementation used Python 3.12.7, Gurobi
13.0.1, NumPy 2.4.3, pandas 2.3.2, SciPy 1.16.0, scikit-learn 1.7.1,
joblib 1.5.2, and Matplotlib 3.10.6. GPU acceleration is not used.
Although the processor provides 24 cores, the main knapsack experiments
used one Gurobi thread per solve.

\section{Additional Experiments}
\label{app:additional_experiments}

We complement the fractional-knapsack experiments with two additional LP cases: solar-grid budget allocation using real installation data and synthetic sparse transportation. Both cases compare the same six methods as the main experiment.

\subsection{Real-Data Solar-Grid Allocation}
\label{app:solar_grid}

\subsubsection{Data and Rolling Evaluation Protocol}
We follow the solar-grid investment setting of \citet{zhou2026credo}, using publicly released monthly aggregates of the real solar-installation data studied by \citet{zhou2025hierarchical}.\footnote{\externalhref{https://github.com/wbzhou2001/CREDO/blob/19e884718599cd390564ddf93aff943a720f3e22/data/real_records.csv}{Monthly solar-installation data released with CREDO}.} The series covers four substations from January 2011--December 2024, with $168$ monthly observations of a four-dimensional count vector and $182$ installations in total. The monthly sequence has no missing observations, and zero-installation months are retained.

For each evaluation year $y\in\{2013,\ldots,2024\}$, we use a $24$-month window from January of year $y-1$ through December of year $y$. The three chronological splits are:
\begin{itemize}
    \item training: January--August of year $y-1$;
    \item calibration: September of year $y-1$--April of year $y$;
    \item testing: May--December of year $y$.
\end{itemize}
Each split contains eight months. Advancing the window by twelve months produces twelve evaluation windows and $96$ distinct test months. The miscoverage level is $\alpha=0.20$ throughout.

Let $N_{tj}$ denote the installation count at station $j$ in month $t$, and let $\mathcal I_y^{\mathrm{tr}}$ be the training months for year $y$. We define the station-specific threshold $\tau_{j,y}$, the smoothed overload-loss coefficient $c_{tj}^{(y)}$, and the sigmoid function $\sigma$ as follows:
\[
\begin{aligned}
\tau_{j,y}&=\frac{1}{8}\sum_{t\in\mathcal I_y^{\mathrm{tr}}}N_{tj},\\
c_{tj}^{(y)}&=\sigma\!\left(0.5(N_{tj}-\tau_{j,y})\right),\\
\sigma(a)&=\frac{1}{1+\exp(-a)}.
\end{aligned}
\]
Thus, each station's threshold is determined by its training-period installation history, and larger installation counts yield larger loss coefficients. The same threshold is used to transform the training, calibration, and test observations within that window.

The point predictor is the historical mean of the transformed training costs:
\[
\hat c_{j,y}=\frac{1}{8}
\sum_{t\in\mathcal I_y^{\mathrm{tr}}}c_{tj}^{(y)}.
\]
This historical-mean setting uses the four station histories directly. Predictions and calibrated sets are held fixed over the eight test months and updated when the window advances.

\subsubsection{Allocation Model and Regret}
Before the monthly loss coefficients are observed, a planner allocates a divisible upgrade budget of two station-equivalents. An upgrade fraction $u_j\in[0,1]$ reduces station $j$'s exposure by the same fraction, subject to $\sum_j u_j\leq2$. Writing $x_j=1-u_j$ for the remaining exposure gives the linear allocation model
\begin{equation}
\begin{aligned}
&\min_{x\in\mathcal X_{\mathrm{sol}}}c_t^\top x,\\
&\mathcal X_{\mathrm{sol}}=\{x\in[0,1]^4:\mathbf1^\top x\geq2\}.
\end{aligned}
\label{eq:solar_allocation}
\end{equation}
Here the window superscript on $c_t$ is suppressed. The objective is a dimensionless smoothed overload loss, and realized regret is
\[
R_t(x)=c_t^\top x-\min_{v\in\mathcal X_{\mathrm{sol}}}c_t^\top v.
\]
The feasible region has the eleven vertices
\[
\mathcal V_{\mathrm{sol}}
=\{v\in\{0,1\}^4:\mathbf1^\top v\geq2\}.
\]
Nominal PTO optimizes the predicted loss over this region. Pure RR chooses a vertex, corresponding to full upgrades at selected stations, while Continuous RR permits fractional upgrades across stations.

\subsubsection{Calibration and Refinement}
Each method computes its score on the eight calibration months using the fixed training-period predictor. The conformal order-statistic index is
\[
k_\alpha=\left\lceil(8+1)(1-0.20)\right\rceil=8,
\]
so the threshold is the largest of that method's eight calibration scores. Pure and Continuous RR share the same SPO+ threshold and uncertainty set.

The Residual-$\ell_2$, Box, and KNN--Ellipsoid baselines follow Appendix~\ref{app:main_performance_setup}. For this historical-mean setting, their context input is constant and the calibrated sets are used over $\mathbb R^4$. KNN uses all eight training observations, $k=\min(100,n_{\mathrm{train}})=8$; its small-sample covariance uses an eigenvalue floor of $10^{-12}$. The score definitions and other baseline parameters are unchanged.

The eleven vertices are explicitly available, so we retain every active SPO+ piece identified by the screening rule. A fixed-piece LP evaluates the worst-case regret of each candidate decision. Pure RR compares all eleven feasible vertices. Continuous RR combines the dual representations of all active pieces into the finite LP of Proposition~\ref{prop:continuous_master}, and its returned decision is evaluated again using the primal fixed-piece oracles. The master objective and recomputed worst-case regret agree within $10^{-6}$. Solver settings follow Appendix~\ref{app:optimization_settings}.

Hindsight optimizer ties are resolved by minimizing predicted cost among the tied vertices, followed by a fixed vertex order. Among equally good Pure-RR candidates, the implementation first minimizes the predicted objective gap and then uses the same fixed order.

\subsubsection{Decision Performance}
All methods are evaluated on the same $96$ test months. We pool their monthly regrets to compute the mean, linearly interpolated Q95, and empirical CVaR95 using the definitions in Appendix~\ref{app:calibration_evaluation}. For this sample size, the upper $5\%$ corresponds to $4.8$ observations, so CVaR95 uses the four largest regrets and $0.8$ of the fifth-largest regret, divided by $4.8$.

Pure and Continuous RR are run consecutively and share the piece models constructed in the Pure stage; the table reports their separately recorded stage times.

In Table~\ref{tab:solar_performance}, Continuous RR reduces CVaR95 from $0.3718$ to $0.2464$, a $33.7\%$ reduction, with a $1.7\%$ increase in mean regret. It also lowers Q95 from $0.2606$ to $0.2059$ and attains the lowest tail regret among the six methods. Residual-$\ell_2$ RO produces a similar tail profile, while Pure RR preserves the nominal decisions and their mean regret.

In all $96$ test months, Continuous RR spreads the budget across all four stations, with station-level upgrade fractions ranging from $0.493$ to $0.507$. Nominal PTO and Pure RR fully upgrade two stations. The application therefore gives a direct allocation interpretation of the mean--tail trade-off.

\begin{table}[!htb]
\centering
\footnotesize
\setlength{\tabcolsep}{2.5pt}
\caption{Decision performance for real-data solar-grid allocation ($\alpha=0.20$, $96$ test months). Change is the decision-change rate; Time is the mean recorded solver-stage time. Lower regret is better.}
\label{tab:solar_performance}
\begin{tabular}{@{}lrrrrr@{}}
\toprule
Method & Mean & Q95 & CVaR95 & Change & Time (s) \\
\midrule
Nominal PTO & \textbf{0.07869} & 0.2606 & 0.3718 & 0.000 & 0.00043 \\
Pure RR & \textbf{0.07869} & 0.2606 & 0.3718 & 0.000 & 0.01193 \\
Continuous RR & 0.08003 & \textbf{0.2059} & \textbf{0.2464} & 1.000 & 0.00547 \\
Residual-$\ell_2$ RO & 0.08006 & 0.2062 & 0.2471 & 1.000 & 0.00140 \\
Box RO & 0.1177 & 0.3541 & 0.4488 & 1.000 & 0.00038 \\
KNN RO & 0.09353 & 0.2365 & 0.3075 & 1.000 & 0.00130 \\
\bottomrule
\end{tabular}
\end{table}

\subsection{Sparse Transportation Linear Programs}
\label{app:transportation}

\subsubsection{Network and Data Generation}
The network has eight origins, eight destinations, and $32$ permitted shipping arcs, half of the $64$ possible origin--destination connections. We use five independent seeds, $42$--$46$. For each seed, supply and demand vectors are drawn independently from $\operatorname{Dirichlet}(2,\ldots,2)$ and then fixed, each summing to one. Draws are screened to keep distinct proper supply and demand subset sums more than $10^{-9}$ apart, giving nondegenerate transportation vertices.

To construct a feasible sparse network, we first solve the complete-network transportation problem with arc costs sampled independently from $\operatorname{Unif}[0,1]$. We retain the resulting spanning-tree support of fifteen arcs and add seventeen arcs sampled without replacement from the remaining connections. The resulting arc set, supply vector, and demand vector are shared by all training, calibration, and test observations for that seed.

We use the quadratic multiplicative-noise generator from Appendix~\ref{app:knapsack_data_generation}. For each arc $e$, let $\theta_e$ denote its feature-coefficient vector, $\mu_e(Z)$ its conditional mean cost, and $\xi_e$ its multiplicative noise:
\[
\begin{aligned}
Z&\sim\operatorname{Unif}([0,4]^{15}),\\
\mu_e(Z)&=(\theta_e^\top Z)^2,\\
C_e&=\mu_e(Z)\xi_e,\qquad
\xi_e\overset{\mathrm{iid}}{\sim}\operatorname{Unif}[0.8,1.2].
\end{aligned}
\]
For the first thirteen features, the coefficients satisfy
\[
\begin{aligned}
\theta_{e\ell}&=D_{e\ell}A_{e\ell},\\
D_{e\ell}&\sim\operatorname{Bernoulli}(0.5),\\
A_{e\ell}&\sim\operatorname{Unif}[0.8,1.2],
\end{aligned}
\]
independently across arcs and features; the final two coefficients are zero. The signal matrix is generated once per seed and held fixed across the three data splits. Contexts and multiplicative noises are independent across observations. Costs are nonnegative and are used directly as minimization coefficients.

\subsubsection{Transportation Model}
Let $\mathcal E$ be the set of permitted arcs. For $e=(i,j)\in\mathcal E$, let $x_{ij}$ denote the shipment from origin $i$ to destination $j$ and $c_{ij}$ its realized unit cost. With supplies $a_i$ and demands $b_j$, the transportation LP is:
\begin{equation}
\begin{aligned}
\min_{x\geq0}\quad &\sum_{(i,j)\in\mathcal E}c_{ij}x_{ij},\\
\text{s.t.}\quad
&\sum_{j:(i,j)\in\mathcal E}x_{ij}=a_i,\quad i=1,\ldots,8,\\
&\sum_{i:(i,j)\in\mathcal E}x_{ij}=b_j,\quad j=1,\ldots,8.
\end{aligned}
\label{eq:transportation_lp}
\end{equation}
Total shipment volume is one. Removing one redundant demand equality gives the standard form
\[
\mathcal X_{\mathrm{tr}}=\{x\in\mathbb R_+^{32}:Ax=h\},
\qquad
A\in\mathbb R^{15\times32},
\]
where $A$ is the balance-constraint matrix of rank $15$ and $h\in\mathbb R^{15}$ contains the eight supplies and seven retained demands. The feasible region has dimension $17$. Each nondegenerate vertex has fifteen positive arcs forming a spanning tree. Pure RR selects such a vertex; Continuous RR allows any feasible shipment plan. For a realized arc-cost vector $c$, regret is
\[
R(x;c)=c^\top x-\min_{v\in\mathcal X_{\mathrm{tr}}}c^\top v.
\]

\subsubsection{Prediction and Calibration}
Each seed uses $5000$ training observations, $2000$ calibration observations, and $500$ test observations. All six methods share these observations and sample splits. The predictor is multi-output polynomial kernel ridge regression with the kernel, tuning grid, and training-only five-fold selection procedure of Appendix~\ref{app:main_performance_setup}. Feature and target standardization are fitted within each training fold. All five seeds select degree $2$ and $\lambda=2.154$, after which the model is refitted on all $5000$ training observations. Predicted costs are returned to the original scale and clipped below at zero.

We use $\alpha=0.10$. Each score has its own threshold, given by order statistic
\[
k_\alpha=\left\lceil(2000+1)(1-0.10)\right\rceil=1801
\]
of its calibration scores. Both SPO+ refinements use the same SPO+ scores and threshold. The Residual-$\ell_2$, Box, and KNN--Ellipsoid baselines follow Appendix~\ref{app:main_performance_setup}, with the known support $c\geq0$ replacing the knapsack sign constraint. Their score definitions, training procedures, and hyperparameters, including $k=100$ for KNN, are unchanged.

\begin{table}[!htb]
\centering
\footnotesize
\setlength{\tabcolsep}{2.5pt}
\caption{Decision performance on the $8\times8$ transportation network with $32$ arcs ($\alpha=0.10$). Entries average per-seed summaries over five seeds, each with $500$ test observations. Change is the decision-change rate; Time is mean online time.}
\label{tab:transportation_performance}
\begin{tabular}{@{}lrrrrr@{}}
\toprule
Method & Mean & Q95 & CVaR95 & Change & Time (s) \\
\midrule
Nominal PTO & \textbf{0.843} & 3.720 & 5.379 & 0.000 & 0.00202 \\
Pure RR & 0.851 & 3.726 & 5.380 & 0.070 & 0.15305 \\
Continuous RR & 1.235 & \textbf{3.466} & \textbf{4.875} & 0.997 & 0.54145 \\
Residual-$\ell_2$ RO & 3.154 & 6.813 & 7.967 & 0.963 & 0.00378 \\
Box RO & 0.913 & 4.066 & 5.661 & 0.240 & 0.00190 \\
KNN RO & 3.235 & 9.609 & 12.874 & 0.851 & 0.00777 \\
\bottomrule
\end{tabular}
\end{table}

\subsubsection{Active-Piece Generation and Refinement}
Active pieces are generated from the nominal transportation vertex in nondecreasing predicted-gap order. The adjacency oracle adds a nonbasic arc to the current spanning tree, follows the resulting cycle, and adjusts alternating flows until one tree arc leaves the support. Positive-arc supports identify distinct vertices. As in the knapsack experiment, only vertices satisfying the active-piece screening condition are expanded.

The generator uses requested prefix budgets
\[
M\in\{1,5,10,25,50,100,200\},
\]
extending through a complete equal-gap plateau when needed. Each retained piece supplies a fixed-piece LP adversary with $c\geq0$. Continuous RR uses an LP master with scenario cuts. Pure RR uses the same cuts and binary arc-selection variables $y_{ij}$ to represent a transportation vertex:
\[
\begin{aligned}
0\leq x_{ij}&\leq\min(a_i,b_j)y_{ij},\\
\sum_{(i,j)\in\mathcal E}y_{ij}&\leq15,\qquad
y_{ij}\in\{0,1\},
\end{aligned}
\]
together with the balance constraints in \eqref{eq:transportation_lp}. Thus Pure RR ranges over all transportation vertices and Continuous RR over the full feasible polytope; the prefix restricts the adversarial pieces.

Let $\mathrm{LB}$ denote the master lower bound and $\mathrm{UB}$ the full-problem upper bound obtained by combining the adversarial upper bound with the omitted-piece bound in Corollary~\ref{cor:gap-ordered-truncation}. Refinement stops when $\mathrm{UB}-\mathrm{LB}$ is at most
\[
10^{-6}\max\{1,|\mathrm{LB}|,|\mathrm{UB}|\},
\]
or the final prefix budget is reached. The constraint-generation loops have per-instance time budgets of $30$ seconds for Pure RR and $20$ seconds for Continuous RR. Solver settings follow Appendix~\ref{app:optimization_settings}, with feasibility and integrality tolerances of $10^{-9}$ for the pure master. Nominal and hindsight LPs use deterministic dual simplex and a fixed secondary objective to resolve optimizer ties.

At the final budget, Pure RR is certified on all $2500$ test instances and Continuous RR on $2493$ of $2500$; neither method reaches its time budget. The performance summaries include the returned feasible decisions for all $500$ test instances per seed.

\subsubsection{Decision Performance}
For each seed, we compute mean regret, Q95, CVaR95, the decision-change rate, and mean online time over its $500$ test observations. Table~\ref{tab:transportation_performance} reports the arithmetic mean of these per-seed summaries over the five seeds. Time includes amortized prediction and baseline-query costs together with optimization, excluding offline training and calibration. Pure and Continuous RR run consecutively within each generation round, sharing active-piece search and model construction; their reported times include this common work and their respective optimization stages.

Continuous RR reduces Q95 from $3.720$ to $3.466$ and CVaR95 from $5.379$ to $4.875$, a $9.4\%$ reduction in CVaR95, while mean regret increases from $0.843$ to $1.235$. Pure RR remains close to nominal average performance and has the lowest mean regret among the robust methods. It changes $6.96\%$ of nominal shipment plans, illustrating selective intervention under coupled supply and demand constraints.

\end{document}